\documentclass[10pt,journal,compsoc]{IEEEtran}

\ifCLASSOPTIONcompsoc
  \usepackage[nocompress]{cite}
\else
  \usepackage{cite}
\fi

\ifCLASSINFOpdf
\else
\fi

\usepackage{makecell}
\usepackage{multirow}
\usepackage{enumitem}
\usepackage{graphicx}
\usepackage{ulem}
\usepackage{subfigure}

\usepackage{url}
\usepackage{enumerate}
\usepackage{booktabs}
\usepackage{multirow}
\usepackage{xspace}
\usepackage{amsmath}
\usepackage{color}
\usepackage{threeparttable}
\usepackage[ruled,linesnumbered]{algorithm2e}
\usepackage{algorithm2e}
\usepackage{amsmath} 
\usepackage{amsfonts} 
\usepackage{threeparttable}
\usepackage{dcolumn}

\newtheorem{definition}{Definition}
\newtheorem{proof}{Proof}

\newtheorem{proposition}{Proposition}

\newcolumntype{d}[1]{D{.}{.}{#1}}% or D{.}{,}{#1} or D{.}{\cdot}{#1}

\newcommand{\eat}[1]{}

\let\oldhat\hat
\renewcommand{\vec}[1]{\mathbf{#1}}
\renewcommand{\hat}[1]{\oldhat{\mathbf{#1}}}

\newcommand{\eg}{\textit{e.g.,}\xspace}

\newcommand{\ie}{\textit{i.e.,}\xspace}

\begin{document}

\title{Heterogeneous Graph Neural Network with Multi-view Representation Learning}

\author{
  Zezhi Shao, Yongjun Xu, Wei Wei, Fei Wang, Zhao Zhang, Feida Zhu
  % Michael~Shell,~\IEEEmembership{Member,~IEEE,}
% \IEEEcompsocitemizethanks{
% \IEEEcompsocthanksitem 

% E-mail: see http://www.michaelshell.org/contact.html
% \IEEEcompsocthanksitem }
\thanks{
Zezhi Shao, Yongjun Xu, Fei Wang and Zhao Zhang are with the Institute of Computing Technology, CAS, Beijing 100190, China. 
Zezhi Shao is also with the University of Chinese Academy of Sciences, Beijing 100049, China.
(e-mail: shaozezhi19b@ict.ac.cn; xjy@ict.ac.cn; wangfei@ict.ac.cn; zhangzhao2021@ict.ac.cn)
}
\thanks{
Wei Wei is with the School of Computer Science and Technology, Huazhong University of Science and Technology, Hubei 430074, China
(e-mail: weiw@hust.edu.cn).
}
\thanks{
Feida Zhu is with the School of Information Systems, Singapore Management University, 178902 Singapore
(e-mail: fdzhu@smu.edu.sg).
}
\thanks{Corresponding Authour: Wei Wei, weiw@hust.edu.cn and Fei Wang, wangfei@ict.ac.cn.}
% \thanks{Manuscript received xxx xx, 2021; revised xxx xx, 2021.}
}

\markboth{Journal of \LaTeX\ Class Files,~Vol.~14, No.~8, August~2015}%
{Shell \MakeLowercase{\textit{et al.}}: Bare Demo of IEEEtran.cls for Computer Society Journals}

\IEEEtitleabstractindextext{%

\begin{abstract}
In recent years, graph neural networks (GNNs)-based methods have been widely adopted for heterogeneous graph~(HG) embedding, due to their power in effectively encoding rich information from a HG into the low-dimensional node embeddings.
However, previous works usually easily fail to fully leverage the inherent heterogeneity and rich semantics contained in the complex local structures of HGs.
On the one hand, most of the existing methods either inadequately model the local structure under specific semantics, or neglect the heterogeneity when aggregating information from the local structure. 
On the other hand, representations from multiple semantics are not comprehensively integrated to obtain node embeddings with versatility.
To address the problem, we propose a \textit{Heterogeneous Graph Neural Network} for HG embedding \textit{within a Multi-View representation learning framework}~(named MV-HetGNN), which consists of a view-specific ego graph encoder and auto multi-view fusion layer.
MV-HetGNN thoroughly learns complex heterogeneity and semantics in the local structure to generate comprehensive and versatile node representations for HGs. 
Extensive experiments on three real-world HG datasets demonstrate the significant superiority of our proposed MV-HetGNN compared to the state-of-the-art baselines in various downstream tasks, \eg node classification, node clustering, and link prediction.

\end{abstract}

}

\maketitle

\IEEEdisplaynontitleabstractindextext

\IEEEpeerreviewmaketitle

\section{Introduction}

% Graph data and the key idea of Graph Neural Networks
\IEEEPARstart{G}{raph} structured data is ubiquitous in the real world,
such as social networks~\cite{social1,social2,ying2018graph, wang2020m2grl} and citation networks~\cite{graphsage, gcn}.
In recent years, \textit{graph neural networks}~(GNNs) have become one of the standard paradigms for analyzing graph structured data. 
The core idea of GNNs is to explore the multi-hop local structure of the target node, \ie the \textit{ego network} or \textit{ego graph}~\cite{idgnn}, 
by stacking multiple layers.
%  to obtain the node representation.
As a basic graph structure, homogeneous graphs consist of only one type of nodes and edges, 
and the ego graph in it has a clear definition and intuition~\cite{2020methods}, \eg \textit{first}-order or \textit{second}-order structure.  
Therefore, GNNs have achieved great success on homogeneous graphs~\cite{gcn,graphsage}. 

\begin{figure}[t]
  \centering
  \includegraphics[width=1\linewidth]{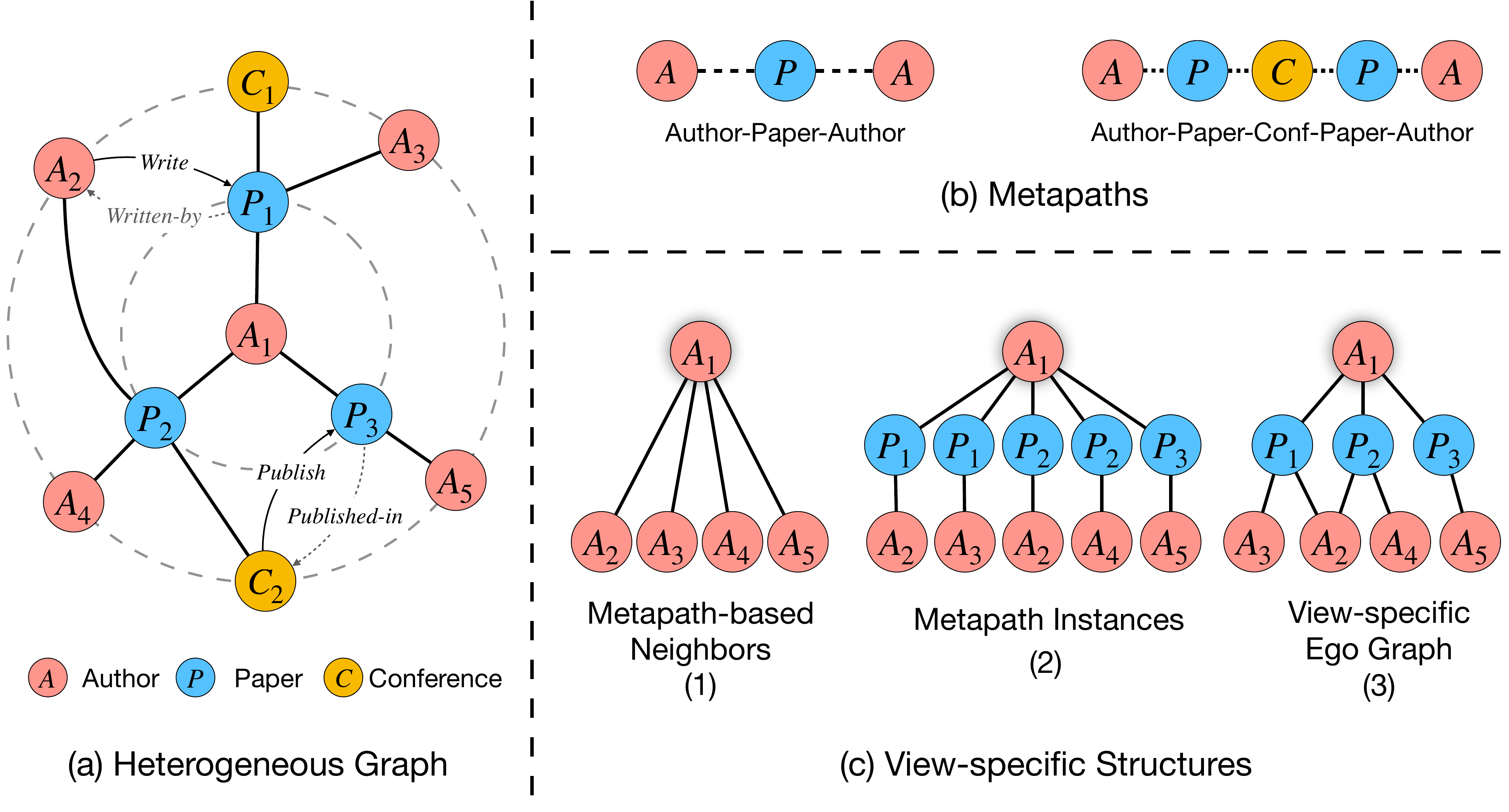}

  \caption{
  An illustrative example of heterogeneous graph~(DBLP) and some key concepts.
  (a) The local structure of node $A_1$ in the heterogeneous graph DBLP.
  (b) Two metapaths in DBLP.
  (c) Three types of metapath-based local structures of author node $A_1$ (based on metapath $APA$) proposed by HAN\cite{HAN}, MAGNN\cite{MAGNN}, and our MV-HetGNN.}
  \label{HetG}
\end{figure}

% The two basic properties of heterogeneous graphs
% However, traditional GNN-based approaches cannot be directly applicable to heterogeneous graphs (HGs) because of two essential properties of HGs: heterogeneity and semantics.
% However, traditional GNN-based methods cannot be directly applied to heterogeneous graphs (HGs) because of two fundamental properties of HGs: heterogeneity and semantics.

However,
traditional GNN-based approaches cannot be directly applied to \textit{heterogeneous graphs}~(HGs)
because of two essential properties of HGs: \textbf{heterogeneity} and \textbf{semantics}.
An example of an ego graph depicting the complex local structure in HGs is shown in Figure \ref{HetG}(a).
\textit{First}, the ego graph of HGs is equipped with multiple types of nodes and relations, \ie \textbf{heterogeneity}.
The features of different types of nodes fall in different feature spaces, hindering the aggregation operation of GNNs.
For example, the \textit{Author} nodes need to be projected into the feature space of the \textit{Paper} nodes through the \textit{Write} relation so as to aggregate with other \textit{Paper} nodes.
\textit{Second}, many meaningful and complex semantic information implicitly exists in the ego graph of HGs, \ie \textbf{semantics}.
These implicit but important relationships are captured by high-order relations, \ie \textit{metapaths}.
Different metapaths reveal different semantics, which can be regarded as a view to observe the target node's local structure. 
For example, as shown in Figure \ref{HetG}(b), \textit{Author-Paper-Author~(APA)} indicates the co-author relationship while the \textit{Author-Paper-Conference-Paper-Author~(APCPA)} indicates the co-conference relationship. 
% These multiple semantic information need to be comprehensively encoded to get versatile node embeddings.

In order to apply GNNs on HGs, a number of works have emerged recently.
Considering the two properties mentioned above, we divide related works into two categories. 
The first category mainly focuses on heterogeneity. 
These works model type-specific mapping functions to project heterogeneous nodes to the same feature space to eliminate heterogeneity.
Then, they apply GNNs and stack multiple layers to catch high-order information~\cite{RGCN,HetGNN,RelGNN,HGT}.
However, these methods do not explicitly make use of semantics.
Moreover, simply stacking multiple layers to catch high-order information causes lower-order semantic information to suffer from over-smooth due to the different lengths of semantics in HGs.
Instead, the second category explicitly utilizes metapaths as semantics\footnote{Following \cite{HAN, MAGNN}, we use semantic and metapath interchangeably in this paper.} to guide information aggregation.
First, they decompose the multi-hop local structure of HGs, which contains complex semantic information, into multiple local structures under different semantics. 
Then, the node representations under different semantics are extracted by aggregating the local structure.
% in a way similar to the first category. 
Finally, representations under different semantics are fused together.
Benefiting from modeling the heterogeneity and semantics simultaneously, these methods are effective and popular~\cite{HAN, MAGNN, 2020WWWR1, 2020SCCR2, MEIRec, RecoGCN, HGSRec}.

{\color{black}
Despite the encouraging results, 
previous works still fail to fully leverage the inherent heterogeneity and rich semantics contained in the complex local structures of HGs.
% these methods still fail to fully explore the heterogeneity and semantics in the complex local structures in HGs.
% 
Specifically, they still suffer from at least one of the following limitations.
(1) The modeling of local structures under each semantics is inadequately~\cite{HAN, MAGNN, HGSRec}.
For example, as shown in Figure \ref{HetG}(c), HAN~\cite{HAN} aggregates information from metapath-based neighborhoods, discarding the intermediate nodes.
MAGNN~\cite{MAGNN} aggregates information at the metapath instance level~(\ie sequence level), 
and neglects the overall graph structure of the local structure, which causes information loss and inflexibility in the aggregation process.
(2) The modeling of the mapping function between heterogeneous nodes is neglected when aggregating the local structures under each semantics~\cite{MEIRec,RecoGCN}. 
That is, the heterogeneity of HGs is not fully explored.
Most of these works only project the feature of heterogeneous nodes to the same dimension\cite{HAN,Tree,HGSRec} and aggregate features directly, neglecting the mapping relation between them.
% 
% (3) The interaction characteristics between multiple semantic information are not comprehensively utilized to obtain versatile representations. 
(3) The representations from multiple semantics are not comprehensively utilized to obtain versatile node representations.
Most methods use simple concatenation or the attention mechanism to fuse representations from different semantics~\cite{HAN, MAGNN, Tree, 2020WWWR1}. 
However, the simple concatenation preserves redundant information among multiple semantics, leading to poor performance.
Meanwhile, the attention mechanism can not theoretically guarantee versatile node embeddings, \ie superior to any single view representation. 
Moreover, our experiments in Sections \ref{Section5} demonstrate that the attention mechanism can not consistently outperform the simple mean average or single view.}
% Similar results can also be found in a very recent work~\cite{2021GIAM}.}

% Our Model
To tackle these issues, 
we model the HG embedding from the perspective of \textit{Multi-view Representation Learning}~(MvRL).
The intuitive idea of introducing MvRL is that the semantic property of HGs conforms to the idea of MvRL, 
  \ie each semantics can be regarded as a view to observe the target node's local structure.
{\color{black}
Different from existing works, the MvRL idea specifically requires adequate modeling of the structure and heterogeneity under each view, and comprehensive integration of representations from multiple views.
On the one hand, within each view, it requires adequately aggregating local structures and modeling heterogeneity to obtain view-specific representations.
On the other hand, among multiple views, it requires comprehensively integrating representations from multiple views into a latent representation with theoretical guarantees of versatility~(\ie superior to any single view) and better performance in downstream tasks.}
% 
% Based on the above motivation, we propose a \textit{Heterogeneous Graph Neural Network based on Multi-view Representation Learning}~(MV-HetGNN) for HG embedding.
% 
Based on the above motivation, we propose a \textit{Heterogeneous Graph Neural Network} for HG embedding \textit{within a Multi-View representation learning framework}~(named MV-HetGNN).
MV-HetGNN solves the above problems with two key components, the view-specific ego graph encoder and auto multi-view fusion layer.
{\color{black}
On the one hand, MV-HetGNN first decomposes the original complete ego graph in HGs into a set of view-specific ego graphs, which preserve complete local structures under each semantics.
Then, the view-specific ego graph encoder leverages TransE~\cite{TransE} to learn the representations of relations in HGs, so as to model the mapping function between heterogeneous nodes and aggregate information at the graph level in a bottom-up manner.
On the other hand, the auto multi-view fusion layer aims to integrate the embeddings from different views comprehensively.
There are two main challenges. 
First, there is much redundant information among multi-view representations~\cite{AE2}, \ie the overlap of representations.
Second, the fused representations require a theoretical guarantee of versatility~\cite{CPMNets}, \ie performing at least as good as any single-view representation.
Auto multi-view fusion layer overcomes both challenges by using hierarchical autoencoders with orthogonal regularization.
In general, our method solves many issues and extends the second category by introducing the idea of MvRL.}
In summary, this work makes several major contributions:
\begin{itemize}
\item
We model the local structure under each semantics as view-specific ego graphs and propose a novel view-specific ego graph encoder module.
It learns the representations of relations to model the mapping function between heterogeneous nodes to address heterogeneity and aggregates information comprehensively.
\item
We propose a novel auto multi-view fusion layer to comprehensively integrate the embeddings from different semantics.
It utilizes hierarchical autoencoders with orthogonal regularization to remove the redundant information cross multiple views and obtain versatile node embeddings with theoretical guarantees.
%   which comprehensively integrates the embeddings from different views
%   to obtain more versatile node embeddings with theoretical guarantee.
% Moreover, we design orthogonal regularization for it.
\item
We conduct extensive experiments on three public real-world HG datasets with three different tasks. 
The experimental results demonstrate the superiority of MV-HetGNN over seven state-of-the-art models.

\end{itemize}

The paper is organized as follows.
Section \ref{Section2} covers related work, and Section \ref{Section3} presents preliminaries and the problem definition.
In Section \ref{Section4}, we present the MV-HetGNN model in detail.
In Section \ref{Section5}, we conduct extensive experiments and visualizations to evaluate the effectiveness of MV-HetGNN.
In particular, Section \ref{Section52} compares the performance of MV-HetGNN with other baselines.
Section \ref{Section_Study_VEGE} compares our view-specific ego graph encoder with existing approaches~\cite{HAN, MAGNN}, and Section \ref{Section_Study_AMF} compares our auto multi-view fusion layer with other popular approaches in detail.
Finally, Section \ref{Section6} concludes the paper.

% To validate our motivations and the disadvantages of existing works.

% In Section 4, we detail the MVHetGNN model. In Section 5, we conduct extensive experiments and visualizations to evaluate the effectiveness of MV-HetGNN. In particular, Section 5.2 compares the performance of MV-HetGNN with other baselines. Section 5.3 compares our view-specific self-graph encoding with existing methods [9], [10] in detail, while Section 5.4 compares our automatic multi-view fusion with other popular methods in detail. Section 6 summarizes the paper.

\section{Related Work}
\label{Section2}
\subsection{Graph Neural Networks}
% GNN的起源：将深度学习和图结构数据结合起来的尝试
% GNN的分类：频域和空域GNN
% 频谱GNN及其缺点：在图上进行图的傅立叶变换以分析其频域；计算困难、没有泛化能力。
% 空域GNN的核心思想：聚合目标节点的局部结构，即Ego Graph，中的信息。
% 举例：GAT、ID-GNN。他们的区别就在于如何设计消息传递函数和聚合函数以抽取Ego Graph中的信息。
% 缺点：没有考虑HG的基本性质，不能直接被用在HG上。

\textit{Graph Neural Networks} (GNNs) are proposed to apply deep neural networks to deal with graph-structured data. 
GNNs can be divided into two categories: spectral-based and spatial-based methods. 
Spectral-based GNNs define convolution operations in the Fourier domain by computing the eigendecomposition of graph Laplacian~\cite{firstgcn,chebnet,gcn}. 
Therefore, these models usually require an entire graph as input, and the eigendecomposition is computationally expensive, making them inefficient and lacking generalization ability. 

Spatial-based GNNs~\cite{graphsage,gat} define convolution operations directly in the graph domain by aggregating the information from the target nodes’ local structures~\cite{mpnn}, \ie ego graph~\cite{idgnn}.
For example, GAT~\cite{gat} aggregates information according to the importance score of neighborhoods assigned by a masked self-attention layer~\cite{transformer}.
Recently, many new powerful GNN models and theories~\cite{amgcn,bestgcn,wltest} have been proposed, which improve the representation power of GNNs. 
In order to exceed the expressive power of the 1-WL test, 
ID-GNN~\cite{idgnn} assigns different message-passing parameters to the central node and other nodes when aggregating information from the ego graph.
{\color{black} Due to their high efficiency, strong performance, and generalization ability, spatial GNNs have become mainstream.
}
However, these powerful models and theories are built for homogeneous graphs, in which the local structure is well defined and has good intuition, e.g., first-order, second-order~\cite{2020methods}. 
The unique properties of HGs, \textbf{heterogeneity} and \textbf{semantics}, are not considered.
Hence, they cannot be directly adapted to HGs. 

\subsection{Heterogeneous Graph Neural Networks}
\label{related_works:HetGNN}
% HetGNN的目的和特点：将GNN扩展到异质图上以全面地利用丰富的节点特征以及语义信息
% HetGNN的分类：依据HG的特点，将方法分为两类。
% 第一类的思想、例子、缺点：
    % 第一类专注于建模Ego Graph中的异质性以应用GNN。解决异质性的方法，是建模映射函数将不同类型的节点映射到同一个特征空间下。
    % 举例：
        % HGT利用node/edge type dependent的注意力机制来处理图异质性。
        % HetSANN利用edge-specific的转换操作来投射把在源节点中的hidden state映射到目标节点类型中。
        % CompGCN使用组合操作（例如TransE、DistMult），来以参数高效地方式学习边的表征以建模映射函数。
    % 缺点：异质图中的语义性没有被充分利用
        % 这些方法通过简单堆叠多层来获取语义信息。
        % 但由于语义信息的长度是不一致的（例如DBLP中的APA、APCPA两个语义信息），简单地堆叠会导致较低阶的语义信息面临过平滑的问题。
% 第二类的思想、例子、缺点
    % 第二类专注于显式地利用语义性。他们利用MetaPath来探索Ego Graph中丰富的语义。这类方法一般分为两步。
    % 首先：利用多个metapath将复杂的Ego Graph降解为多个特定于语义的局部结构，并编码其中的信息。
    % 举例：
        % HAN/HGSRec利用MetaPath将局部结构建模为metapath-based neighborhood，以获取高阶信息。但它忽略了中间节点以避免处理异质性。
        % MAGNN在MetaPath Instance Level，即Sequence Level上编码信息。但它忽略了局部结构的图结构。
        % 也有一些与本工作建模方式相似的方法，例如MEIRec、RecoGCN、T-GCN。然而，MEIRec忽略了MetaPath中间节点和目标节点的信息，且他们都忽略了异质节点之间映射关系的建模。
    % 第二：需要利用探索多种语义以得到更好的节点表征。
    % 举例：
        % HAN/MAGNN/T-GCN/HGSRec/RecoGCN都利用Attention-based的方法来选择更重要的语义；
    % 缺点：
        %  但是一方面，Attention-based方法多种语义信息之间的角度特性没有被充分利用，得到的节点表征无法保证完备性，即保证比单个MetaPath表现更好。
        % 另一方面，有文献GIAM表明Attention-based的方法容易引起过拟合。
% 总的来说，相比于现存工作，一方面我们详尽地建模了特定语义下的局部结构，并解决了局部结构中的异质性以充分地聚合信息。另一方面，我们综合编码所有语义以得到完备的节点表征，得到了更好的效果。

%  ------------------------- Heterogeneous Graph Neural Networks -------------------------
Heterogeneous GNN models~\cite{GATNE,zhang2019shne,yd1,fu2019metapath,yd2,zhu2019relation,RGCN,HGT,HetSANN,HAN,MAGNN,Tree} extend GNN techniques on HGs to fully use the rich node features and semantic information. 
According to the two basic properties of HGs, GNNs can be divided into two categories.

%  -------------------------      First Category      -------------------------
The first category focuses on modelling the heterogeneity of the ego graph. 
These models eliminate heterogeneity by modelling mapping functions to project heterogeneous nodes to the same feature space to apply GNNs.
For example, HGT~\cite{HGT} introduces the node- and edge-type dependent attention mechanism to handle graph heterogeneity.
HetSANN~\cite{HetSANN} uses edge-type specific transformation operations to project the hidden state in the space of source node type to the hidden space of the target node type.
% Graph attention mechanism~\cite{gat} are use to aggregate information.
CompGCN~\cite{comgcn} uses composition operations, such as TransE~\cite{TransE} or DistMult~\cite{DistMult}, to model the mapping function efficiently by learning edge representation vectors.
Although they try to thoroughly model the heterogeneity to apply GNNs, the semantics property of HGs are not fully utilized.
% 这些方法简单地堆叠多层来获取高阶语义信息。然而由于语义信息的阶数是不一致的，例如APA、APCPA，简单的叠加多层信息会引起较低阶的信息过平滑。
Moreover, since the order of semantic information is inconsistent, \eg \textit{APA} and \textit{APCPA} in the DBLP dataset, 
simply stacking multiple layers to catch high-order semantic information will cause lower-order semantic information to face the over-smooth problem.

%  -------------------------      Second Category      -------------------------
The second category focuses on semantics.
These models explicitly utilize semantic information (captured by meta-paths) to analyze the ego graph.
Specifically, these methods are generally divided into two steps.
\textit{First}, the ego graph is decomposed by multiple metapaths to get local structures, which are further encoded to get node representations under each semantics.
For example, as shown in Figure \ref{HetG}(c) HAN~\cite{HAN} and HGSRec~\cite{HGSRec} model the local structure under each semantics as metapath-based neighbours.
However, they discard all intermediate nodes along metapath.
MAGNN~\cite{MAGNN} fixes that by aggregating information at the metapath instance level, \ie sequence level.
However, the graph structure of local structures is neglected, which causes information loss and inflexibility in the aggregation process.
Similar to our model, several works leverage graph-like structures to model the local structure under each semantics, such as MEIRec~\cite{MEIRec}, RecoGCN~\cite{RecoGCN}, T-GCN\cite{Tree}.
However, MEIRec neglects the feature information of intermediate and target nodes. 
In addition, they all ignore the modelling of mapping functions between heterogeneous nodes.
\textit{Second}, diverse semantics need to be exploited to get superior representations.
Attention-based methods are widely used in these models~\cite{HAN,MAGNN,Tree,HGSRec,RecoGCN} to softly select the most meaningful metapath and fuse the representations under different semantics.
However, 
% the interactive characteristics between multiple semantic information have not been fully utilized and
the final representations cannot be guaranteed theoretically superior to any single semantic representation.

%  -------------------------      In general      -------------------------
In general, although the heterogeneous graph neural networks have made notable progress, these problems still hinders the performance of heterogeneous graph neural networks, and there is still much room for improvement by solving these problems.

\section{Problem Definition}

\begin{table}
    \caption{Important notations used in this paper.}
    \label{tab:notations}
    \begin{tabular}{m{1.3cm}|m{6.3cm}}
      \toprule
      \textbf{Notations} & \textbf{Definitions}\\
      \midrule
      % \midrule
      $\mathcal{V}$ & The set of nodes in a graph \\
      $\mathcal{A}$ & The set of node types \\
      $\mathcal{E}$ & The set of edges in a graph \\
      $\mathcal{R}$ & The set of edge types, \ie relations \\
      $\mathcal{G}$ & A graph $\mathcal{G}=\{\mathcal{V, E}\}$\\
      $P$ & A metapath \\
      $\mathcal{P}$ & The set of metapaths\\
      $v$ & A node $v\in\mathcal{V}$\\
      $r$ & A kind of relation $r\in \mathcal{R}$ \\
      $\mathcal{N}_v$ & The set of neighbors of node $v$\\
      $\mathcal{EG}_{v}^{P}$ & The ego graph of node $v$ under metapath $P$\\
      $\mathbf{h}_v$ & Hidden state of node $v$\\
      $\mathbf{h}_r$ & Hidden state of relation $r$\\
      $\mathbf{H}$ & Hidden states of nodes with the same node type\\
      $\odot$ & The hadamard product \\
      $|\cdot|$ & The cardinality of a set\\
      $||\cdot||_F$ & Frobenius norm of a matrix \\
      $||\cdot||_1$ & L1 norm of a matrix \\
      \bottomrule
    \end{tabular}
  \end{table}

\label{Section3}
% 给出一些重要的术语的定义，使用图片的形式对一些定义进行展示，并在一个表格中对所有符号进行总结。
\begin{definition}
\label{def-1}
\textbf{Heterogeneous Graph.} 
A heterogeneous graph $\mathcal{G}=(\mathcal{V},\mathcal{E}; \mathcal{\phi}, \mathcal{\omega})$ 
    is composed of a vertex set $\mathcal{V}$ and an edge set $\mathcal{E}$, 
    along with object type mapping function $\phi: \mathcal{V} \rightarrow \mathcal{A}$ 
    and edge type mapping function $\omega: \mathcal{E}\rightarrow \mathcal{R}$. 
    $\mathcal{A}$ and $\mathcal{R}$ denote the predefined sets of object types and edge types, respectively, 
        where $|\mathcal{A}| + |\mathcal{R}|>2$.
% 要不要删掉这句
$\mathcal{R}$ could be further split into two subsets: $\mathcal{R}^+$ and $\mathcal{R}^-$. 
Note that $self\ loop$ relation can be randomly included in one of the two categories.
An example is given in Figure \ref{HetG}(a).
\end{definition}

\begin{definition}
\label{def-2}
\textbf{Metapath.}
Consider $A_i \in \mathcal{A}$ and $R_i \in \mathcal{R}$ denote a node type and an edge type, respectively, 
    a metapath $P$ is defined as a path in the form of $A_1 \stackrel{R_1}{\longrightarrow} A_2 \stackrel{R_2}{\longrightarrow} \cdots \stackrel{R_l}{\longrightarrow} A_{l+1}$, 
    which describes a composite relation $R=R_1 \circ R_2 \circ \cdots \circ R_l$ between object types $A_1$ and $A_{l+1}$, 
    where $\circ$ denotes the composition operator over relations.
Examples are given in Figure \ref{HetG}(b).
\end{definition}

\begin{definition}
\label{def-3}
\textbf{Metapath based Ego Graph.}
Given a metapath $P: A_1 \rightarrow A_2\rightarrow...\rightarrow A_N$ and a target node $v$ with type $A_N$,
the ego graph ($\mathcal{EG}^{P}_v$) is a directed graph induced by the metapath-based neighborhoods and intermediate nodes along metapaths as well as $v$ itself.
An example is shown in Figure \ref{HetG}(c).
\end{definition}

\begin{definition}
\label{def-4}
\noindent \textbf{Heterogeneous Graph Embedding.}
% Given a heterogeneous graph $\mathcal{G=(V,E; \phi, \psi)}$ with node attribute matrices $\mathbf{X}_{A_i}\in \mathbb{R}^{|\mathcal{V}_{A_i}|\times d_{A_i}}$ for node types $A_i\in \mathcal{A}$, heterogeneous graph embedding aims to learn the lower-dimensional node representations $\mathbf{h}_v\in \mathbb{R}^d$ for all $v\in \mathcal{V}$, where $d\ll |\mathcal{V}|$.
Heterogeneous graph embedding aims to learn a function that embeds the nodes in $\mathcal{G}=(\mathcal{V},\mathcal{E}; \mathcal{\phi}, \mathcal{\omega})$ into a d-dimensional Euclidean space where $d\ll |\mathcal{V}|$.
% while preserving rich structural and semantic information involved in $\mathcal{G}$.
\end{definition}

\begin{definition}
\label{def-5}
\noindent \textbf{Versatility of Multi-View Representations.~\cite{CPMNets}}
Given representations $\mathbf{h}^{1},\cdots,\mathbf{h}^{V}$ from $V$ views, 
the multi-view representation $\mathbf{h}$ is of versatility if $\forall$ $v$ and $\forall$ mapping $\varphi(\cdot)$ with $y^{v}=\varphi(\mathbf{h}^{v})$, 
there exists a mapping $\psi(\cdot)$ satisfying $y^v=\psi(\mathbf{h})$.
\end{definition}
\section{The Proposed Framework}
\begin{figure*}[ht]
  \centering
  \includegraphics[width=1\linewidth]{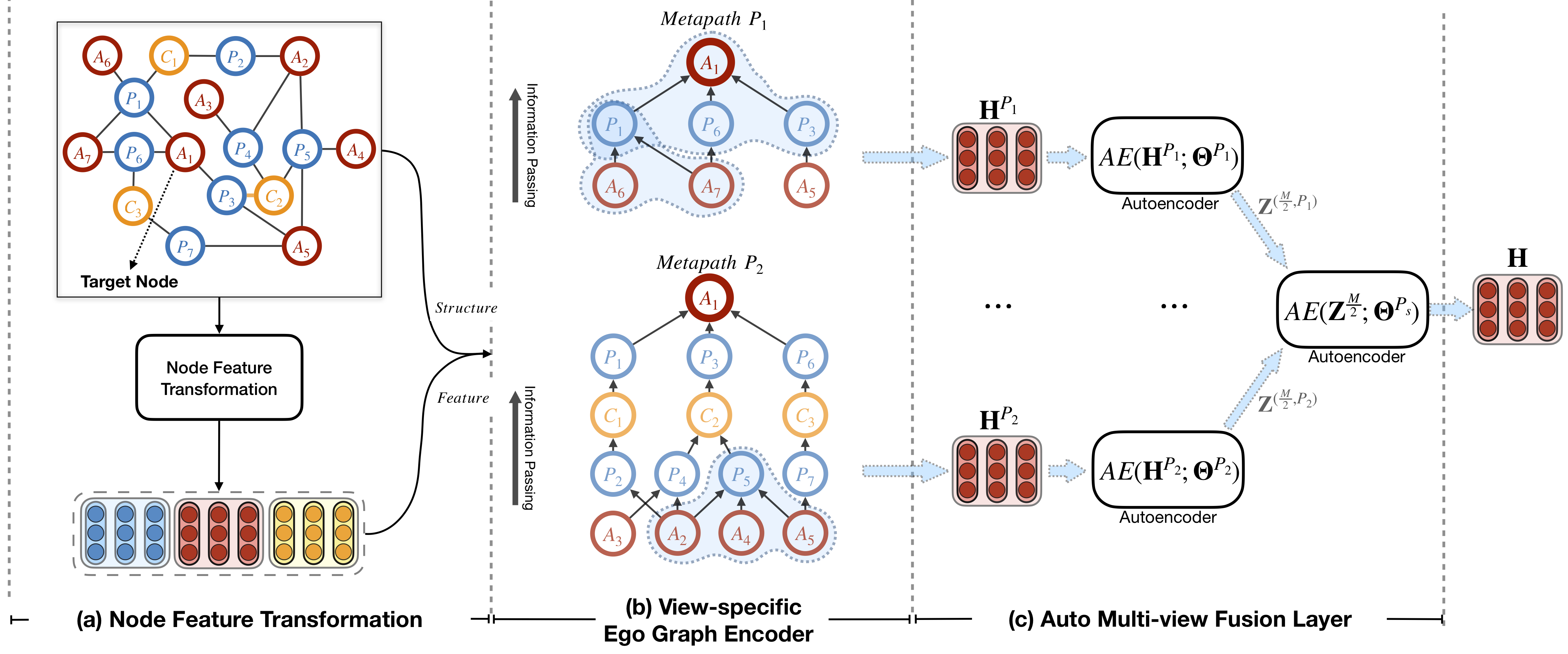}
  \caption{
  The overall architecture of MV-HetGNN. 
  (a) Node feature transformation projects the feature vectors of all types of nodes to the same dimension. 
  (b) View-specific ego graph encoder generates embedding for a target node under each view, such as $A_1$ in the picture above.
  (c) Auto multi-view fusion layer integrates multi-view embeddings by hierarchical autoencoders to obtain more versatile node embeddings.
  }
  \label{LGGNN}
\end{figure*}

{\color{black}
\label{Section4}
In this section, we present the MV-HetGNN, which sufficiently models the heterogeneity and semantics in the complex local structure in HGs from the perspective of multi-view representation learning.
As shown in Figure \ref{LGGNN}, MV-HetGNN contains three primary steps.
First, since different node types are associated with features of different dimensions, 
  the node feature transformation converts them into features of the same dimension.
Second, we treat each semantic~(captured by the metapaths) $P$ as the view to observe target nodes' local structure
  and conduct view-specific ego graph encoder to obtain the node embeddings under each view.
Finally, the auto multi-view fusion layer 
  comprehensively integrates the embeddings from different views to obtain versatile node embeddings.
}

\subsection{Node Feature Transformation}
% 目的/方法
Feature vectors with different dimensions are troublesome when aggregating information in subsequent modules. 
To address this issue, 
    we apply node-type specific transformations to 
    convert heterogeneous feature vectors into features of the same dimension.
Given any node type $A_i\in \mathcal{A}$, the node feature transformation layer can be shown as follows:
\begin{equation}
\mathbf{H}_{A_i}^{'}=\sigma(\mathbf{X}_{A_i}\mathbf{W}_{A_i}), 
\label{eq:fc}
\end{equation}
where 
$\mathbf{H}_{A_i}^{'}\in \mathbb{R}^{|\mathcal{V}_{A_i}|\times d^{'}}$ is the transformed latent vectors of nodes of $A_i$ type.
$\mathbf{X}_{A_i}\in \mathbb{R}^{|\mathcal{V}_{A_i}|\times d_{A_i}}$ is the original feature vector of all $A_i$ nodes. 
$\mathbf{W}_{A_i}\in \mathbb{R}^{d_{A_i}\times d^{'}}$ is the type-specific parameters, and $d^{'}$ is the unified dimension. 
$\sigma$ means the activation function, such as ReLU~\cite{ReLU}. 

The node feature transformation unifies the feature dimensions and facilitates the usage of subsequent modules.
However, it should be noted that the heterogeneity has not been resolved.
Although the dimensions of the feature vector of all nodes are the same, they still lie in different embedding spaces, so they should not be aggregated directly.

% 需要添加View的说法
\subsection{View-specific Ego Graph Encoder}
% 异质图的局部结构，即Ego Graph非常复杂，每一层内有多类节点，层间有多种关系，且蕴含着多种语义信息。
% 在本方法中，我们利用多个语义信息将完整的Ego Graph进行降解为多个基于MetaPath的Ego Graph。
% 这样操作的好处是，可以将多种语义信息解耦。
% 并且基于MetaPath的Ego Graph有着良好的特性：每一层内仅有一种节点，层间仅有一种关系，仅代表一种语义。
% 对该Metapath based Ego Graph进行编码，即可得到节点在该View下的表征。
% 需要注意的是在编码过程中需要充分地解决异质性的问题。
Considering the \textbf{heterogeneity} and \textbf{semantics} of HGs, 
  the local structure of the target node in HGs is rather complex. 
{\color{black}
This module aims to model the local structure under each semantics adequately and aggregate information with addressing the heterogeneity.}

Specifically, we treat each semantic $P$ as the view to observe the target node’s local structure.
{\color{black}Specifically, the original complete ego graph is} decomposed into multiple view-specific ego graphs.
The advantage of this operation is that multiple semantics can be decoupled.
Furthermore, compared to metapath-based neighborhoods~\cite{HAN} and metapath instances~\cite{MAGNN}, the view-specific ego graph preserves more complete local structure under each semantics.
% compared with the metapath-based neighborhood~\cite{HAN} and the metapath instance~\cite{MAGNN}.
Another benefit is that the view-specific ego graph has better characteristics: 
  there is only one type of node in the same order, 
  and there is only one type of relation between different orders.
Given a metapath $P=A_1 \stackrel{R_1}{\longrightarrow} A_2 \stackrel{R_2}{\longrightarrow} \cdots \stackrel{R_{N-1}}{\longrightarrow} A_{N}$ 
  and a target node $v$ with $\phi(v)=A_N$, 
the view-specific ego graph $\mathcal{EG}_v^{P}$ preserves the original and complete multi-hop structure 
  under semantic $P$, where the nodes of the $i$th hop have the same node type $A_{N-i}$ and the relation between the $i$th hop and the $(i-1)$th hop~($i\geq 1$) is $R_{N-i}$. 
{\color{black}
An example is shown in Figure \ref{HetG}(c).}

\begin{figure}
  \centering
  \includegraphics[width=1\linewidth]{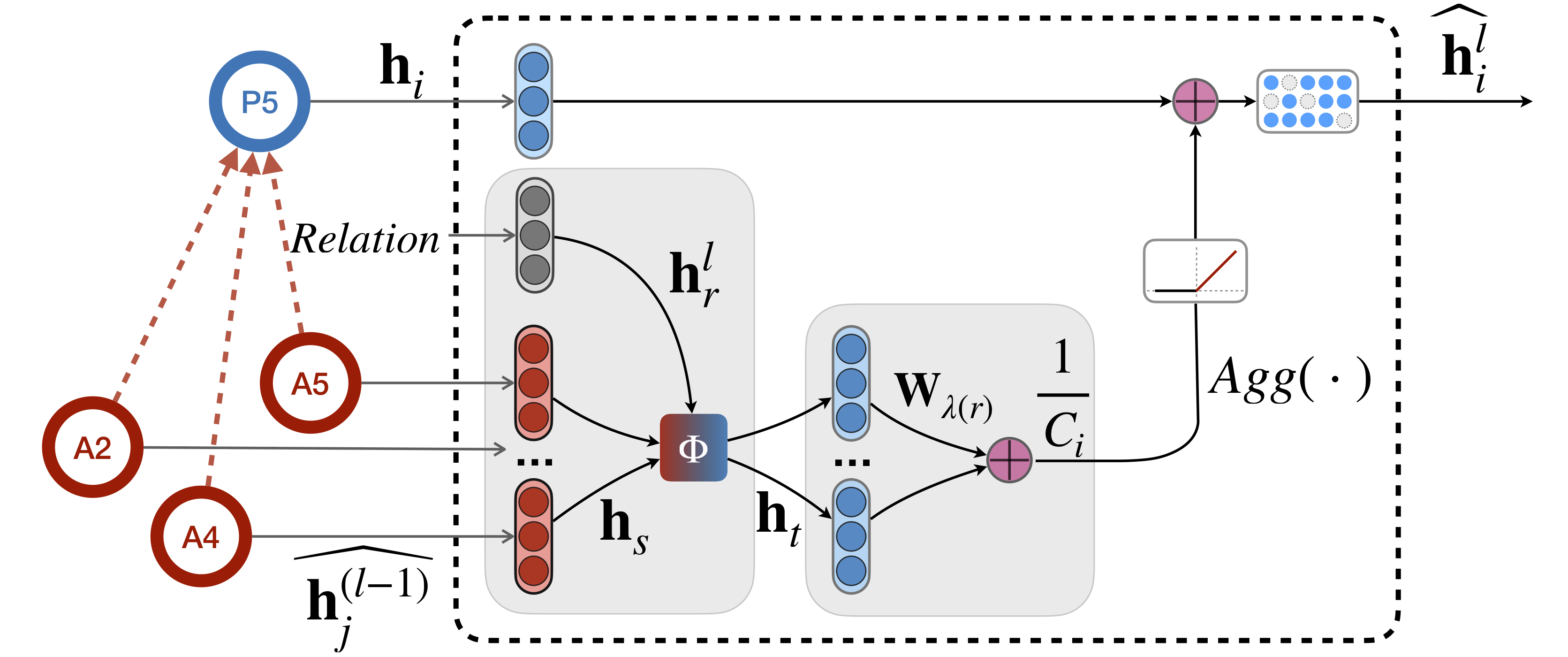}
  \caption{Information aggregation for node $P5$ at level $l$. The Author node's feature vector ($\mathbf{h}_s$), relation vector ($\mathbf{h}_r^{(l)}$), and $P5$'s feature vector ($\mathbf{h}_i$) are aggregated to obtain new encoded feature vector for $P5$. 
  % is projected to the Paper feature space through $\Phi(\mathbf{h}_s, \mathbf{h}_r)$, and aggregated by message passing operation. After an activation function, the Author nodes' information is combined with the hidden features of $P5$ to get the final features. 
%   An activation function is then applied to the encoded neighborhood information, followed by residual connection to get the updated node embedding.
  %An activation function is then applied, followed by residual connection. 
}
  \label{fig:overall_HGE}
\end{figure}

% ego graph encoder
Further, in order to handle the heterogeneity and encode the structural and feature information adequately from the ego graph, 
  we develop an ego graph encoder, which has two advantages.
First, it can aggregate node features guided by the structure of the view-specific ego graph. 
{\color{black}
Second, it handles the heterogeneity in the ego graph by modeling the representations of relations and the mapping function between heterogeneous nodes.}
Two example of view-specific ego graph encoder is illustrated in Figure \ref{LGGNN}(b).
Assuming that level $1$ is the bottom level and level $K$ is the top level,
  ego graph encoder aggregates information from level $1$ to $K$, 
    updating the representation $\mathbf{h}_i^{(l)}\in \mathbb{R}^{d'}$ of each node $i$ at level $l$. 
The aggregation process~(blue region in Figure \ref{LGGNN}(b)) 
  between any intermediate two levels is illustrated in Figure \ref{fig:overall_HGE}.
Specifically, we set the hidden state of node $i$ at level $1$ as $\mathbf{h}_i^{(1)}=\mathbf{h}_i^{'}$, which is generated by the node feature transformation module. 
For any specific node $i$ at level $l$ ($l\ne 1$), we first aggregate its neighborhood information and then apply an activation function. Combined with its own features, the encoded feature is calculated as
\begin{equation}
\mathbf{h}_i^{(l)}=o(\mathbf{h}_i^{'} + \sigma(\mathbf{Agg}(\{\mathbf{h}_j^{(l-1)}, v_j\in \mathcal{N}_i^{(l-1)}\}, \mathbf{h}_r^l))),
\label{res}
\end{equation}
% 结果
where $\mathbf{h}_i^{(l)}\in \mathbb{R}^{d'}$ is the encoded hidden vector of node $i$ at level $l$, 
% 残差
and $\mathbf{h}_i^{'}\in \mathbb{R}^{d'}$ is its initial state generated by the component of node content transformation. 
% 邻居
$\mathcal{N}_i^{(l-1)}$ is the neighborhood set of node $i$ at level $(l-1)$, 
$\mathbf{h}_j^{(l-1)}\in \mathbb{R}^{d'}$ is the encoded hidden vector of node $j$ at level $(l-1)$, 
$\mathbf{h}_r^l$ is the representation of the relation $r$ between level $l$ and $(l-1)$. 
$\sigma(\cdot)$ is the ReLU activation function. 
Note that the representation is randomly initialized and optimized jointly with the network parameters.
$o$ is the dropout layer. 
$\mathbf{Agg}(\cdot)$ is an aggregate function to capture 1-hop neighborhoods information and relation representation: 
\begin{equation}
  \begin{split}
&\mathbf{Agg}(\{\mathbf{h}_j^{(l-1)}, v_j\in \mathcal{N}_i^{(l-1)}\}, \mathbf{h}_r^l)\\
&=\frac{1}{C_i}\sum_{v_j\in \mathcal{N}_i^{(l-1)}} \mathbf{W}_{\lambda(r)}\Phi(\mathbf{h}_j^{(l-1)}, \mathbf{h}_r^l),
  \end{split}
  \label{eq:agg}
\end{equation}
where 
$\Phi(\cdot, \cdot)$ is the mapping function for handling the heterogeneity,
$\mathbf{W}_{\lambda(r)} \in \mathbb{R}^{d'\times d'}$ is the relation-specific message passing parameter,
{\color{black}
$C_i$ is the normalization term.
For simplicity, we set $C_i=|\mathcal{N}_i^{(l-1)}|$, \ie mean average, in our method.
Kindly note that with the benefit of the graph structure, we can further improve the performance by setting $C_i$ to be learnable, \eg through attention mechanisms. 
Related experiments can be found in Section \ref{Section_Study_VEGE}}.

{\color{black} For one thing}
, benefiting from the unified dimensions of all types of nodes, we can model the relations $\mathbf{h}_r^{(l)}$ as $\mathbb{R}^{d^{'}}$ vectors and use the knowledge graph embedding approach to model the mapping function between the feature vectors of different types of nodes. 
{\color{black} Many functions can be adopted here and we apply TransE~\cite{TransE}, which gains an impressive performance and efficiency in our experiments.}
In TransE~\cite{TransE}, for a relation triplet $(s, r, t)$, there will be $\vec{s} + \vec{r} \approx \vec{t}$. Therefore:
\begin{equation}
\Phi(\mathbf{h}_s,\mathbf{h}_r) = \mathbf{h}_s + \mathbf{h}_r.
\end{equation}

{\color{black}For another thing}, the message passing parameter $\mathbf{W}_{\lambda(r)}$ would suffer from the over-parameterization problem with the growth of the number of relations, since each relation $r$ is associated with a matrix $\mathbf{W}_{\lambda(r)}$. Inspired by~\cite{direction,comgcn}, we simplify it to a direction-specific matrix, i.e., $\lambda(r)=dir(r)$, which is defined as follows: 
\begin{equation}
\label{eq:direction}
\mathbf{W}_{\lambda(r)}=\mathbf{W}_{dir(r)}=\left\{ 
\begin{aligned} 
& \mathbf{W}_O, \quad r\in \mathcal{R}^+  \\ 
& \mathbf{W}_I, \quad r\in \mathcal{R}^- 
\end{aligned} 
\right. 
,
\end{equation}
where $\mathcal{R}^+$ and $\mathcal{R}^-$ are opposite sets of two relations. For example, $\mathcal{R}^+=\{write, publish\}$ and $\mathcal{R}^-=\{written-by, published-in\}$. Note that the \textit{self loop} relation can be randomly included in one of the two categories.

After $K-1$ times of calculation on ego graph $\mathcal{EG}_v^{P}$, the representation of the target node $v$ is $\mathbf{h}_v^{(K)}$. 
% Note that the length of metapath is inconsistent. 
Note that the lengths of the metapaths are inconsistent, 
  which results in different times of aggregation of Eq. (\ref{res}).
\begin{equation}
\mathbf{h}_v^{P}= \frac{\mathbf{h}_v^{(K)}}{depth(\mathcal{EG}_v^{P})}
\label{HGE:out}
\end{equation}
where $depth(\mathcal{EG}_v^{P})$ is the depth of the view-specific ego graph, which equals to the length of $P$. 

In summary, 
given the features generated by the node feature transformation 
  and the view set $\mathcal{P}_{A_i}=\{P_1,\cdots,P_{|\mathcal{P}_{A_i}|}\}$
  where the metapath start or end with the node type $A_i\in\mathcal{A}$, the view-specific ego graph encoder will generate a set of representations 
    under each view for node $v\in\mathcal{A}_i$, 
  denoted as $\{\mathbf{h}_v^{P_1}, \mathbf{h}_v^{P_2}, \ldots, \mathbf{h}_v^{P_{|\mathcal{P}_{A_i}|}}\}$.

% TODO Guided by metapath
\subsection{Auto Multi-view Fusion Layer}
% 然后，我们面临另一个更关键的问题，如何融合多视角下的表征，即一个多视角融合问题。
% After the view-specific ego graph encoder, 
%     we get diverse representations of target nodes from each view.
% Then, there is another critical problem of fusing representations from multiple views, \ie the multi-view fusion problem.
{\color{black}
In this section, we aim to fuse diverse representations from multiple views.
% 我们主要面临着两个挑战：
There are two key challenges.
% 需要保证完备性。完备性很重要，代表了包含全部的信息量。以往的方法，例如Mean Pooling或Attention based problem[citations]，无法保证完备性。
First, the fused representation requires a theoretical guarantee of a crucial characteristic, versatility~\cite{CPMNets}~(Definition \ref{def-5}), which means the fused representation can perform at least as good as any single-view representation. 
Popular approaches in existing HG embedding works, such as attention-based approaches~\cite{HAN, MAGNN, 2020WWWR1, 2020SCCR2}, can not hold the versatility.
% 需要去除多视角之间的冗余信息。一直观的保留所有信息的方法是Concatenation，然而，它保留了节点之间非常多的冗余信息，即overlapping。
Second, there is much redundant information among multi-view representations, which must be removed in the fusion process~\cite{AE2}.
The simple concatenation operation preserves the redundant information, leading to poor performance, even if it holds the versatility intuitively.
% 为了解决这两个挑战，我们提出了Auto Multi-View Fusion Layer，他由一个带正交正则的层次化AutoEncoder构成。
The auto multi-view fusion layer addresses these two challenges through hierarchical autoencoders and orthogonal regularization.}

\subsubsection{Hierarhical Autoencoders}

Given a metapath $P_j$ in view set $\mathcal{P}_{A_i}$, 
    we denote the embeddings of all nodes generated by view-specific ego graph encoder as $\mathbf{H}^{P_j}=[\mathbf{h}^{P_j}_1, \mathbf{h}^{P_j}_2, \cdots, \mathbf{h}^{P_j}_{|\mathcal{V}_{A_i}|}]^T\in\mathbb{R}^{|\mathcal{V}_{A_i}|\times d^{'}}$.
The representations from all views $\{\mathbf{H}^{P_1}, \mathbf{H}^{P_2}, \cdots, \mathbf{H}^{P_{|\mathcal{P}_{A_i}|}}\}$ need be encoded into versatile representations $\mathbf{H}$.
{\color{black}
In order to do that, hierarchical autoencoders are used to preserve intra-view information and encode inter-view information simultaneously.}
We first demonstrate a standard autoencoder as $AE(\mathbf{X};\mathbf{\Theta})$, 
where $\mathbf{X}$ is the input of autoencoder and $\mathbf{\Theta}=\{\mathbf{W}_{ae}^{(m)}, \mathbf{b}_{ae}^{(m)}\}|_{m=1}^M$ is the network parameters with $M$ being the number of layers.
The first $\frac{M}{2}$ layers are the encoder network, 
    and the last $\frac{M}{2}$ layers are the decoder network. 
% For view $P_j$, let $\mathbf{Z}^{(0, P_j)}=\mathbf{H}^{P_j}$, then the output of the $m$th layer is:
% \begin{equation}
% \mathbf{Z}^{(m, P_j)}=\sigma(\mathbf{Z}^{(m-1,P_j)}\mathbf{W}_{ae}^{(m,P_j)}+\mathbf{b}_{ae}^{(m,P_j)}),
% \end{equation} 
% where $\mathbf{Z}^{(m, P_j)}\in\mathbb{R}^{|\mathcal{V}_{A_i}|\times d^{m}}$. 
% $d_m$ is the output dimension of $m$-th layer. 
% $\mathbf{W}_{ae}^{(m,P_j)}\in\mathbb{R}^{d^{m-1}\times d^{m}}$ and $\mathbf{b}_{ae}^{(m, P_j)}\in\mathbb{R}^{d^m}$ denote the weights and bias of the $m$th layer, respectively.
Let $\mathbf{Z}^{(0)}=\mathbf{X}$, then the output of the $m$th layer is:
\begin{equation}
\mathbf{Z}^{(m)}=\sigma(\mathbf{Z}^{(m-1)}\mathbf{W}_{ae}^{(m)}+\mathbf{b}_{ae}^{(m)}),
\end{equation} 
where $\mathbf{Z}^{(m)}\in\mathbb{R}^{\cdot\times d^{m}}$. 
$d_m$ is the output dimension of $m$-th layer. 
$\mathbf{W}_{ae}^{(m)}\in\mathbb{R}^{d^{m-1}\times d^{m}}$ and $\mathbf{b}_{ae}^{(m)}\in\mathbb{R}^{d^m}$ denote the weights and bias of the $m$th layer, respectively.
$\sigma(\cdot)$ is the non-linear activation, such as ReLU~\cite{ReLU}.
For simplicity, we denote $f(\cdot;\mathbf{\Theta}_e)$ as the encoder network where parameters $\mathbf{\Theta}_e=\{\mathbf{W}_{ae}^{(m)}, \mathbf{b}_{ae}^{(m)}\}|_{m=1}^{\frac{M}{2}}$ 
and 
$f(\cdot;\mathbf{\Theta}_d)$ as the decoder network, where parameters $\mathbf{\Theta}_d=\{\mathbf{W}_{ae}^{(m)}, \mathbf{b}_{ae}^{(m)}\}|_{m=\frac{M}{2}}^M$.
$f(\cdot)$ denotes general the multilayer perceptron architecture.

First, in intra-view, we use view-specific autoencoders to further compress the representations of each view to be more compact.
Assuming the autoencoder under view $P_j$ is $AE(\mathbf{H}^{P_j};\mathbf{\Theta}^{P_j})$, 
    then the compressed representations, \ie the output of the encoder network is:
\begin{equation}
    \mathbf{Z}^{(\frac{M}{2},P_j)}=f(\mathbf{H}^{P_j};\mathbf{\Theta}^{P_j}_e).
    \label{intra_en}
\end{equation}
The reconstruction of the decoder network is:
\begin{equation}
    \mathbf{Z}^{(M, P_j)}=f(\mathbf{Z}^{(\frac{M}{2},P_j)};,\mathbf{\Theta}^{P_j}_d).
    \label{intra_de}
\end{equation}
Therefore, the reconstruction loss for all view-specific autoencoders is:
\begin{equation}
    \mathcal{L}^{intra}_{re}=\frac{1}{2}\sum_{P_j}||\mathbf{H}^{P_j}-\mathbf{Z}^{(M, P_j)}||^2_F.
    \label{intra_re}
\end{equation}

Second, in inter-view, we use a 2-layer supervised autoencoder~\cite{SAE} to encode all information into a latent representation comprehensively.
We first concatenate the compressed representations from all views, denoted as $\mathbf{Z}^{\frac{M}{2}}=[\mathbf{Z}^{(\frac{M}{2}, P_1)},\cdots, \mathbf{Z}^{(\frac{M}{2}, \mathcal{P}_{A_i})}]$.
Assuming the supervised autoencoder network is $AE(\mathbf{Z}^{\frac{M}{2}};\mathbf{\Theta}^{s})$, 
where $\Theta^{s}=\{\mathbf{W}_{s}^1,\mathbf{W}_{s}^2,\mathbf{b}_{s}^1,,\mathbf{b}_{s}^2\}$ and 
$\mathbf{W}_{s}^1\in\mathbb{R}^{(d^{\frac{M}{2}}\cdot|\mathcal{P}_{A_i}|)\times d}$ and 
$\mathbf{W}_{s}^2\in\mathbb{R}^{d\times  (d^{\frac{M}{2}}\cdot|\mathcal{P}_{A_i}|)}$ and
$\mathbf{b}^1_{s}\in\mathbb{R}^{d}$ and
$\mathbf{b}^2_{s}\in\mathbb{R}^{d^{\frac{M}{2}}\cdot|\mathcal{P}_{A_i}|}$, then the output of the encoder network is:
\begin{equation}
    \mathbf{H}=f(\mathbf{Z}^{\frac{M}{2}};\mathbf{\Theta}_e^s)=\sigma(\mathbf{Z}^{\frac{M}{2}}\mathbf{W}_{s}^1+\mathbf{b}_{s}^1).
    \label{inter_en}
\end{equation}
The reconstruction of the decoder network is:
\begin{equation}
    \mathbf{Z}^{\frac{M}{2}}_{re}=f(\mathbf{H};\mathbf{\Theta}_d^s)=\sigma(\mathbf{H}\mathbf{W}_{s}^2+\mathbf{b}_{s}^2).
    \label{inter_de}
\end{equation}
Therefore, the reconstruction loss for the supervised autoencoders is:
\begin{equation}
    \mathcal{L}^{inter}_{re}=\frac{1}{2}||\mathbf{Z}^{\frac{M}{2}}_{re}-\mathbf{Z}^{\frac{M}{2}}||^2_F.
    \label{inter_re}
\end{equation}
The supervised information is the downstream task loss and will be discussed in Section \ref{sec:optimization}.

Ideally, minimizing Eq. (\ref{intra_re}) and Eq. (\ref{inter_re}) 
    will offer versatility to the multi-view representations $\mathbf{H}\in\mathbb{R}^{|\mathcal{V}_{A_i}|\times d}$, \ie the final representations of all nodes of type $A_i$.
Here we give a brief theoretical proof:
% Next, we will proof that our final representation $\mathbf{H}$ is complete and can theoretically guarantee at least as good as any single view $P_i$.
\begin{proposition} 
    \textbf{(Versatility of the Multi-View Representation $\mathbf{H}$)} There exists a solution to Eq. (\ref{inter_re}) and Eq. (\ref{intra_re}) which holds the versatility.
\end{proposition}

\begin{proof}
% 我们首先对Decoding矩阵进行分块
% 一次对于任意View P_i，
\textit{
First, we partition the weight parameter matrix of the decoder of the 2-layer supervised autoencoder as:
\begin{equation}
    \mathbf{W}_s^2=[(\mathbf{W}_s^2)^{P_1}, \cdots, (\mathbf{W}_s^2)^{P_{|\mathcal{P}_{A_i}|}}],
\end{equation}
where $(\mathbf{W}_s^2)^{P_j}\in\mathbb{R}^{|\mathcal{V}_{A_i}|\times d^{\frac{M}{2}}}$.
According to Eq. (\ref{inter_re}), 
    it is easy to show that there exists 
    $\mathbf{Z}^{(\frac{M}{2}, P_j)}=
    \sigma(\mathbf{H}(\mathbf{W}_s^2)^{P_j} + \mathbf{b}_s^2)=
    f(\mathbf{H};\mathbf{\Theta}_d^{(s, P_j)})$, 
    where $f(\cdot;\mathbf{\Theta}_d^{(s, P_j)})$ is the mapping from the
        multi-view representation $\mathbf{H}$ to the compressed single view representation $\mathbf{Z}^{(\frac{M}{2}, P_j)}$.
    % where $f(\cdot;\mathbf{\Theta}_d^{(s, P_j)})$ is the mapping from $\mathbf{H}$ to compressed view representation $\mathbf{Z}^{(\frac{M}{2}, P_i)}$.
    % where $f(\cdot)$ 
    % is the mapping from $\mathbf{H}$ to $\mathbf{Z}^{(\frac{M}{2}, P_j)}$.
Further, according to Eq. (\ref{intra_re}), there exists $\mathbf{H}^{P_j}=f(\mathbf{Z}^{(\frac{M}{2}, P_j)}; \mathbf{\Theta}_d^{P_j})$.
Consequently, 
there exists:
\begin{equation}
    \mathbf{H}^{P_j}=
    f(f(\mathbf{H};\mathbf{\Theta}_d^{(s, P_j)});\mathbf{\Theta}_d^{P_j}\})=
    f(\mathbf{H};\{\mathbf{\Theta}_d^{(s, P_j)}, \mathbf{\Theta}_d^{P_j}\}),
\end{equation}
where $f(\cdot;\{\mathbf{\Theta}_d^{(s, P_j)}, \mathbf{\Theta}_d^{P_j}\})$ 
    is the mapping from $\mathbf{H}$ to $\mathbf{H}^{P_j}$.
Hence, $\forall$ $\varphi(\cdot)$ with $\mathbf{Y}^{P_j}=\varphi(\mathbf{H}^{P_j})$, 
there exists a mapping $\psi(\cdot)$ satisfying $\mathbf{Y}^{P_j}=\psi(\mathbf{H})$
by defining $\psi(\cdot)=\varphi(f(\cdot;\{\mathbf{\Theta}_d^{(s, P_j)}, \mathbf{\Theta}_d^{P_j}\}))$.
}

\end{proof}

\subsubsection{Orthogonal Regularization}
\label{Section_Ortho}
{\color{black}There exist two issues when applying hierarchical autoencoders.
First, the representations from different views contain much redundant information as mentioned before. For example, the \textit{Co-author} view and \textit{Co-conference} view may share many paper and author nodes.
Second, the autoencoders may suffer from the over-parameterization problem.
Since the graph datasets are often semi-supervised where only a few labels can be accessed, the optimization of autoencoders might be problematic.
In order to address these issues, 
    we introduce orthogonal regularization to the encoder network of each autoencoder.
On the one hand, the orthogonal column vectors can be regarded as orthogonal bases, ensuring each encoded dimension represents a unique meaning, \ie independent of the other dimensions~\cite{OrthorAE}. 
On the other hand, 
due to the orthogonality constraint, only informative weights are non-zero.}

Specifically, considering all the first $\frac{M}{2}$ layers of all autoencoders $ae$, the orthogonal loss is:
\begin{equation}
    \mathcal{L}_{ortho}=\sum_{ae}\sum_m^{\frac{M}{2}}||(\mathbf{W}_{ae}^{m})^T(\mathbf{W}_{ae}^{m})\odot(\mathbf{1}_{d_m}-\mathbf{I}_{d_m})||_1,
    \label{ortho}
\end{equation}
where $\odot$ refers to the element-wise product.
We do not set orthogonal constraints on the decoder networks 
    because the decoder needs to reconstruct the original hidden state 
    that may contain many redundant features.

\subsection{Optimization}
\label{sec:optimization}
In MV-HetGNN, we adopt a task-guided training strategy to optimize the network parameters. 
There are several strategies to optimize MV-HetGNN.
For example, optimize view-specific ego graph encoder and auto multi-view fusion layer sequentially according to the downstream tasks.
% For 扩展性和便捷性，我们将这些Loss放在一起调整。这样做的好处是本方法就可以被扩展到多中下游任务。
However, for the scalability and flexibility of the model, we optimize the MV-HetGNN in an \textbf{end-to-end} paradigm.

Specifically, according to the node labels' availability of downstream tasks, we divide the them into two categories: 
    semi-supervised and unsupervised downstream tasks. 
For semi-supervised learning~(\eg node classification and node clustering), 
    we first define the labeled node set as $\mathcal{Y}_L$ 
    and the the downstream classifier as $\mathbf{W}_C\in\mathbb{R}^{C\times d}$, where $C$ is the number of classes.
Then the downstream loss is:
\begin{equation}
    \mathcal{L}_{ds}=-\sum_{v\in\mathcal{Y}_{L}} \mathbf{y}_v \ln(\mathbf{W}_C\mathbf{h}_v^T),
\end{equation}
where the $\mathbf{h}_v\in \mathbb{R}^{1\times d}$ and $\mathbf{y}_v\in\mathbb{R}^{1\times C}$ is the embedding vector and one-hot label vector of labeled node $v$.
For unsupervised learning~(\eg link prediction), we design the loss function of preserving the graph structure, \ie the adjacency relationship of nodes. 
We optimize the model parameters by minimizing the following loss function through negative sampling~\cite{word2vec}:
\begin{equation}
\mathcal{L}_{ds}=-\sum_{(u,v)\in \Omega} \log\sigma(\mathbf{h}^T_u\cdot\mathbf{h}_v)-\sum_{(u',v')\in\Omega^-}\log \sigma(-\mathbf{h}_{u'}^T\cdot \mathbf{h}_{v'}),
\label{eq:unsupervised}
\end{equation}
where $\sigma(\cdot)$ is the sigmoid function, $\Omega$ is the set of positive node pairs, and $\Omega^-$ is the sampled set of negative node pairs sampled from all unobserved node pairs.

Therefore, the complete loss function of MV-HetGNN is:
\begin{equation}
    \mathcal{L}=\mathcal{L}_{ds} + \lambda(\mathcal{L}^{intra}_{re} + \mathcal{L}^{inter}_{re}) + \mathcal{L}_{ortho},
    \label{total_loss}
\end{equation}
where $\lambda$ is a critical hyper-parameter that controls the degree of versatility.
It is often a relatively small number, such as 0.1, 0.05.
This is mainly because:
i) 
At the beginning of training, 
    the representation of each view is initialized almost randomly, 
    so the reconstruction loss~(\ie loss of generality) is meaningless.
Therefore, in order to optimize the network parameter in the right direction, 
    the downstream task loss needs to play a major role.
ii) 
In practical cases, it is usually difficult to guarantee the exact versatility. 
Moreover, exactly versatile multi-view representation will loss flexibility for various datasets and downstream tasks.

We optimize the model parameters by minimizing $\mathcal{L}$ via backpropagation and gradient descent.
The overall learning algorithm is shown in Algorithm \ref{alg}.

% 注意和普通的MV Learning不同，
% 每一个MetaPath输出的内容（不是很准确）在一开始的时候都是随机初始化的，
% 因此让Reconstruction Error Loss不能在优化过程中占据过多权重，
% 否则会导致Fusion Layer学到完全没有意义参数，从而使得整个模型Train不起来。

\normalem
\begin{algorithm}[htp]
\renewcommand\arraystretch{1.0}
\caption{The overall learning algorithm of MV-HetGNN}
\LinesNumbered
\label{alg}
\KwIn{ The graph $\mathcal{G}=(\mathcal{V}, \mathcal{E})$, 
the node features $\{\mathbf{H}_{A_i}, \forall A_i\in\mathcal{A}\}$, 
the view~(metapath) set $\mathcal{P}$, 
the ego graphs $\mathcal{EG}^{P}_{v}$ of each node $v$ under each view $P$.
}

\KwOut{The node Embeddings $\mathbf{H}$.}
\For {node type $A_i\in\mathcal{A}$}{
    Node feature transformation: $\mathbf{H}_{A_i}^{'}=\sigma(\mathbf{X}_{A_i}\mathbf{W}_{A_i})$;
}

\For {metapath $P_j$ in $\mathcal{P}$}{
    \For {node $v\in\mathcal{V}$}{
        Given $\mathcal{EG}^{P_j}_v$, calculate the view-specific representation 
            $\mathbf{h}_v^{P_j}$ by the view-specific ego graph encoder;
    }
}
Calculate multi-view representation $\mathbf{H}$ by Eq. (\ref{intra_en}, \ref{inter_en});\\
Calculate the reconstruction loss by Eq. (\ref{intra_re}, \ref{inter_re});\\
Calculate the orthogonal regularization loss by Eq. (\ref{ortho});\\
Backpropagation and update parameters according to Eq. (\ref{total_loss});
\end{algorithm}

\section{Experiments}
\label{Section5}
\subsection{Experimental Setup}
\noindent \textbf{Dataset}. 
As shown in Table \ref{tab:datasets}, three heterogeneous graph datasets from different domains are used to evaluate the performance of MV-HetGNN. Following~\cite{HAN,MAGNN}, we use three widely used evaluation tasks: node classification, node clustering, and link prediction.  

\begin{itemize}
\item \textbf{DBLP} \footnote{https://dblp.uni-trier.de/}: 
We adopt a subset of DBLP extracted by~\cite{DBLP_e}. 
The authors are divided into four areas. 
DBLP is used for node classification and node clustering with data partition of 400~(9.86\%), 400~(9.86\%), and 3257~(80.28\%) for training, validation, and testing. 

\item \textbf{IMDb} \footnote{https://www.imdb.com/}: 
We adopt a subset of IMDb extracted by~\cite{MAGNN}. 
Each movie is labeled as one of three classes. 
IMDb is used for both node classification and node clustering with data partition of 400~(9.35\%), 400~(9.35\%), and 3478~(81.30\%) nodes for training, validation, and testing, respectively. 

\item \textbf{Last.fm} \footnote{https://www.last.fm/}: 
We adopt a subset of Last.fm released by~\cite{LastFM}. 
No label or features are included. 
Last.fm dataset is used for the link prediction task. 
All links are divided into training, validation, and testing set with data partition of 18567~(20\%), 9283~(10\%), and 64984~(70\%), respectively. 

\end{itemize}

\begin{table}
  \caption{Statistics of datasets.}
  \label{tab:datasets}
  \begin{tabular}{c||l|l|c}
    \toprule
    \textbf{Dataset} & \textbf{\# Node} & \textbf{\# Edge} & \textbf{Metapaths}\\
    % \midrule
    \midrule
    \textbf{DBLP} & 
    \makecell[l]{Author(A): 4,057 \\ Paper(P): 2,081 \\ Term(T): 7,723 \\Conference(C): 20} & 
    \makecell[l]{A-P: 19,645 \\ P-T: 85,810 \\ P-C: 14,328} & 
    \makecell[c]{APA \\ APTPA \\ APCPA} \\

    \midrule
    \textbf{IMDb} & 
    \makecell[l]{Movie(M): 4,278 \\ Director(D): 2,081 \\ Actor(A): 5,257} & 
    \makecell[l]{M-D: 4,278 \\ M-A: 12,828} & 
    \makecell[c]{MDM, MAM \\ DMD, DMAMD \\ AMA, AMDMA} \\

    \midrule
    \textbf{Last.fm} & 
    \makecell[l]{User(U): 1,892 \\ Artist(A): 17,632 \\ Tag(T): 1,088} & 
    \makecell[l]{U-U: 12,717 \\ U-A: 92,834 \\ A-T: 23,253} & 
    \makecell[c]{UU, UAU \\ UATAU, AUA \\ AUUA, ATA} \\

    \bottomrule
  \end{tabular}
\end{table}

\begin{table*}[ht]
\renewcommand\arraystretch{1.0}
  \centering
  \caption{Results~(\%) on the DBLP and IMDb datasets for node classification task.}
  \label{tab:node classification}
%   \begin{tabular}{|c|c|c|c|c|c|c|c|c|c|c|}
  \begin{tabular}{ccc|cc|cccccc}
    \toprule
    \multirow{2}*{\textbf{Dataset}} & \multirow{2}*{\textbf{Metrics}} & \multirow{2}*{\textbf{Train\%}} & \multicolumn{2}{c|}{\textbf{Unsupervised}} & \multicolumn{6}{c}{\textbf{Semi-supervised}}\\ 
    \cmidrule(r){4-5} \cmidrule(r){6-11}
        &    &    & \textbf{Metapath2vec} & \textbf{HERec} & \textbf{GCN}& \textbf{GAT}& \textbf{HAN}& \textbf{HGT} & \textbf{MAGNN} & \textbf{MV-HetGNN}\\
    \midrule
    \multirow{8}*{\textbf{IMDb}} & \multirow{4}*{Macro-F1} 
           & 20\%& 45.94& 45.31& 53.63 & 54.74& 57.52& 59.38& \uline{59.48}& \textbf{61.33} \\
        % \cline{3-11}
        &  & 40\% & 47.41& 46.63& 53.86& 56.27& 57.81& \uline{59.91}& 59.79& \textbf{61.43}\\
        % \cline{3-11}
        &  & 60\% & 48.23& 47.07& 54.22 & 56.97& 58.28& \uline{60.32}& 60.02& \textbf{61.39}\\
        % \cline{3-11}
        &  & 80\% & 50.34& 48.02& 54.77& 57.43& 58.69& \uline{60.38}& 60.20& \textbf{61.89}\\

        \cmidrule{2-11}

        & \multirow{4}*{Micro-F1} & 20\% & 47.47& 46.19& 53.61& 54.56& 57.79& \uline{59.42}& 59.27& \textbf{61.31}\\
        % \cline{3-11}
        &  & 40\%  & 48.69& 48.03& 53.88& 56.17& 58.77& \uline{60.08}& 59.92& \textbf{61.43}\\
        % \cline{3-11}
        &  & 60\%  & 49.54& 48.41& 54.19& 56.89& 59.11& \uline{60.27}& 60.14& \textbf{61.36}\\
        % \cline{3-11}
        &  & 80\%  & 50.47& 49.57& 54.12& 57.48& 59.57& \uline{60.44}& 60.21& \textbf{61.89}\\
        % \cline{3-11}

    \midrule
    \multirow{8}*{\textbf{DBLP}} & \multirow{4}*{Macro-F1} & 20\% & 89.39& 90.31& 90.00& 91.37& 91.87& 92.05& \uline{93.01}& \textbf{95.23}\\
        % \cline{3-11}
        &  & 40\%  & 89.99& 91.15& 90.11& 91.70& 92.36& 92.57& \uline{93.24}& \textbf{95.31}\\
        % \cline{3-11}
        &  & 60\%  & 90.31& 92.01& 90.12& 91.76& 92.80& 92.90& \uline{93.52}& \textbf{95.34}\\
        % \cline{3-11}
        &  & 80\%  & 90.94& 92.37& 91.07& 93.81& 93.01& 93.40& \uline{93.79}& \textbf{95.44}\\

        \cmidrule{2-11}

        & \multirow{4}*{Micro-F1} 
           & 20\% & 90.43& 91.49& 90.03& 91.89& 92.48& 92.55& \uline{93.51}& \textbf{95.52}\\
        % \cline{3-11}
        &  & 40\%  & 90.99& 92.05& 90.31& 92.17& 92.90& 93.08& \uline{93.74}& \textbf{95.64}\\
        % \cline{3-11}
        &  & 60\%  & 91.33& 92.66& 90.31& 92.32& 93.35& 93.38& \uline{94.01}& \textbf{95.56}\\
        % \cline{3-11}
        &  & 80\%  & 91.61& 92.78& 90.40& 92.36& 93.53& 93.46& \uline{94.17}& \textbf{95.80}\\
    
    \bottomrule
  \end{tabular}
\end{table*}

\noindent \textbf{Baselines}. 
We compare MV-HetGNN with the following baselines:
\begin{itemize}
\item \textbf{Metapath2vec}~\cite{metapath2vec} uses metapath-guided random walks and  skip-gram model~\cite{skip-gram} to generates node embeddings. 
We test Metapath2vec++~\cite{metapath2vec} on all metapaths separately and report the best results. 
\item \textbf{HERec}~\cite{HERec} applies DeepWalk~\cite{deepwalk} on multiple metapath-based homogeneous graphs, and proposes an fusion algorithm for rating prediction. 
For node classification/clustering, we report the best result of all metapaths. 
\item \textbf{GCN}~\cite{gcn} performs convolutional operations in the graph Fourier domain. 
\item \textbf{GAT}~\cite{gat} conducts convolutional operations in the spatial domain with the attention mechanism. 
We test GCN and GAT on metapath-based homogeneous graphs and report the best results. 
\item \textbf{HAN}~\cite{HAN} learns node embeddings from different metapath-based homogeneous graphs and leverages the attention mechanism to combine them into one vector for each node. 
\item \textbf{HGT}~\cite{HGT} designs node- and edge-type dependent parameters to characterize the heterogeneous attention over each edge to model heterogeneity. 
\item \textbf{MAGNN}~\cite{MAGNN} uses intra-metapath aggregation and inter-metapath aggregation to encode metapath instances, and learn the importance of different metapaths by an attention mechanism. 
\end{itemize}

The above baselines can be divided into four categories. 
Metapath2vec and HERec are the shallow heterogeneous graphs embedding models. 
GCN and GAT are conventional GNNs designed for homogeneous graphs.
HGT is the first category of heterogeneous graph neural networks~(as described in Section \ref{related_works:HetGNN}).
HAN and MAGNN are the second category of heterogeneous graph neural networks.

\noindent \textbf{Implementation}. 
For Metapath2vec and HERec, we set the window size to 5, walk length to 100, walks per node to 40, and the number of negative samples to 5. 
For all GNNs, we set the dropout rate to 0.5 as default. 
For HAN, MAGNN, and HGT, we follow the original setting reported in these papers. 
For MV-HetGNN, we employ the Adam optimizer, 
  and set the learning rate to 0.005, 0.001, 0.001 for DBLP and IMDb and Last.fm dataset, respectively.
We set the number of layers of view-specific auto encoder $M$ as 2.
{\color{black}{
  For the dimension hyper-parameters 
  $d'$~(output feature dimension of node feature transformation),  
  $d^{\frac{M}{2}}$~(the output dimension of the encoder of view-specific autoencoder) and 
  $d$~(the dimension of multi-view representations),
  we set $d'=2d^{\frac{M}{2}}=2d$ for DBLP, 
  and $d'=d^{\frac{M}{2}}=d$ for IMDb and 
  $d'=d^{\frac{M}{2}}=d$ for Last.fm datasets, respectively.
\label{setup}
}}

\subsection{Experimental Results}
\label{Section52}
\noindent \textbf{Node Classification}. 
We conduct node classification experiments~\cite{HAN,MAGNN} on the DBLP and IMDb datasets, with about 10\% data used for training.
After obtaining embeddings of labeled data by each model, we feed the testing nodes into a linear support vector machine~(SVM) classifier with varying training proportions. 
Since the variance of graph-structured data can be relatively high, we repeat it ten times and report the averaged \textit{Macro-F1} and \textit{Micro-F1} in Table \ref{tab:node classification}. 

As shown in the table, 
the bold and underlined numbers indicate the best and runner-up results in the row, respectively. 
MV-HetGNN consistently achieves the best performance. 
The shallow models~(Metapath2vec and HERec) perform worse than GCN or GAT, since they do not leverage node content features. 
HAN obtains better performance than GAT and GCN because it exploits multiple metapaths to explore various semantics. 
HGT and MAGNN are able to outperform HAN because they can utilize more node features by stacking multilayer or metapath encoder, respectively. 
MV-HetGNN consistently outperforms HGT, which stacks multiple layers to catch high-order semantics. 
In contrast, MV-HetGNN employs the view-specific ego graph encoder to utilize higher-order information effectively. 
MV-HetGNN also obtains better results than MAGNN. 
There are two main reasons.
For one thing, MV-HetGNN model the metapath-based local structure more comprehensively than MAGNN.
For another thing, MV-HetGNN can comprehensively integrate the embeddings from different views 
    to obtain more versatile embeddings than MAGNN.
%   to obtain more versatile embeddings while MAGNN use attention only to softly \textit{select} metapaths.

\begin{table*}[ht]
\renewcommand\arraystretch{1.0}
  \centering
  \caption{Results (\%) on the DBLP and IMDb datasets for node clustering task.}
  \label{tab:node clustering}
  \begin{tabular}{cc|cc|cccccc}
    \toprule
    \multirow{2}*{\textbf{Dataset}} &  \multirow{2}*{\textbf{Metrics}} & \multicolumn{2}{c|}{\textbf{Unsupervised}} & \multicolumn{6}{c}{\textbf{Semi-supervised}}\\ 
    \cmidrule(r){3-4} \cmidrule(r){5-10}
        &    & \textbf{metapath2vec} & \textbf{HERec} & \textbf{GCN}& \textbf{GAT}& \textbf{HAN} & \textbf{HGT} & \textbf{MAGNN} & \textbf{MV-HetGNN}\\
    % \hline
    \midrule
    \multirow{2}*{\textbf{IMDb}} & NMI  & 0.93 & 0.41 & 8.10 & 9.98 & 12.46 & \uline{14.50} & 14.41 & \textbf{16.04}\\
    % \cline{2-10}
     & ARI  & 0.32 & 0.17 & 6.62 & 9.02 & 11.21 & \uline{15.92} & 15.22 &\textbf{17.67}\\

    % \hline
    \midrule
    \multirow{2}*{\textbf{DBLP}} & NMI  & 74.09 & 69.31 & 72.75 & 75.03 & 77.86 & 77.47 & \uline{80.52} & \textbf{84.23} \\
    % \cline{2-10}
     & ARI  & 78.32 & 72.71 & 73.13 & 81.73 & 83.23 & 81.84 & \uline{85.68} & \textbf{89.05}\\

    \bottomrule
  \end{tabular}
\end{table*}

\noindent \textbf{Node Clustering}. 
We conduct node clustering experiments on the DBLP and IMDb datasets, using the same setting as~\cite{HAN,MAGNN}.
We feed the embeddings of labeled nodes to a K-Means algorithm. The number of cluster K is set to 3 for IMDb and 4 for DBLP.
Since the clustering result of the K-Means algorithm is highly dependent on the initialization of the centroids, we repeat K-Means 10 times and report the averaged \textit{normalized mutual information} (NMI) and \textit{adjusted Rand index} (ARI). 

The results are reported in Table~\ref{tab:node clustering}.   
Overall, the relative performance of node clustering task is similar to the node classification task. 
MV-HetGNN significantly performs much better than all baselines consistently. 
The experimental results demonstrate that MV-HetGNN is able to learn more effective representation for the nodes of heterogeneous graphs.
We note that the performance of all evaluated models on IMDb is much worse than on DBLP because every movie node has multiple genres in the original IMDb dataset, but only the first one is chosen as its class label.

\begin{table*}[ht]
  \centering
\renewcommand\arraystretch{1.0}
  \caption{Results (\%) on the Last.fm datasets for link prediction task.}
  \label{tab:link prediction}
  \begin{tabular}{cc|cccccccc}
    \toprule
    \textbf{Dataset} & \textbf{Metrics} & \textbf{metapath2vec} & \textbf{HERec} & \textbf{GCN} & \textbf{GAT} & \textbf{HAN} & \textbf{HGT} & \textbf{MAGNN} & \textbf{MV-HetGNN}\\
    % \hline
    \midrule
    \multirow{2}*{\textbf{Last.fm}} 
    & AUC & 74.32 & 73.98 & 76.59 & 80.03 & 81.00 & 86.53 & \uline{87.68} & \textbf{92.80}\\
    % \cline{2-10}
    \cmidrule{2-10}
     & AP & 74.11 & 72.53 & 75.51 & 81.44 & 82.03 & 88.77 & \uline{89.25} & \textbf{94.03}\\

    \bottomrule
  \end{tabular}
\end{table*}

\noindent \textbf{Link Prediction}. 
We evaluate the performance of link prediction task on Last.fm, following the MAGNN model~\cite{MAGNN}. Compared to MAGNN~\cite{MAGNN}, we adopt a lower training ratio (only 20\% links are used for training), which is a more challenging setting. 
The connected user-artist pairs are treated as positive links, while unconnected user-artist pairs are regarded as negative links.

We add the same number of randomly sampled negative node pairs to the validation and testing sets. 
The GNNs are then optimized by minimizing Equation \ref{eq:unsupervised}. Given the user embedding $\mathbf{h}_u$ and the artist embedding $\mathbf{h}_a$ generated by the trained model, the linking probability of $u$ and $a$ is calculated by
$Prob_{ua} = \sigma (\mathbf{h}_u^T \cdot \mathbf{h}_a)$, 
where $\sigma$ is the commonly used sigmoid function. 
The embedding models are evaluated by the \textit{area under the ROC curve (AUC)} and \textit{average precision} (AP) scores. In Table \ref{tab:link prediction}, we report the averaged results of 10 runs of each embedding model. 
% HAN outperforms Metapath2vec, GCN, and GAT, showing that considering multiple metapaths is crucial to generate more comprehensive node embeddings. 
% Compared with HAN, both HGT and MAGNN are able to achieve large improvements. 
MV-HetGNN outperforms all baseline models, indicating the superiority of MV-HetGNN. 
Compared with other methods, MV-HetGNN can effectively obtain high-order information, and comprehensively use the features and semantic information. 

% \color{black}
\begin{table*}
    \centering
\renewcommand\arraystretch{1.0}
\caption{\color{black}Results (\%) of the study on view-specific ego graph encoder and auto multi-view fusion layer on three datasets.}
\label{tab:Ablation Study}
\begin{tabular}{c|cccc|cccc|cc}
    \toprule 
    % \midrule
    \multirow{2}{*}{\textbf{Variants}} & \multicolumn{4}{c|}{\textbf{IMDb}} & \multicolumn{4}{c|}{\textbf{DBLP}} & \multicolumn{2}{c}{\textbf{Last.fm}}\\
    \cmidrule(r){2-5} \cmidrule(r){6-9} \cmidrule(r){10-11}
     & \textbf{Macro-F1} & \textbf{Micro-F1} & \textbf{NMI} & \textbf{ARI} & \textbf{Macro-F1} & \textbf{Micro-F1} & \textbf{NMI} & \textbf{ARI} & \textbf{AUC} & \textbf{AP}\tabularnewline
    \midrule 
    \textbf{MV-HetGNN} & \uline{61.51} & \uline{61.50} & \textbf{16.04} & \textbf{17.67} & \uline{95.33} & \uline{95.63} & \textbf{84.23} & \textbf{89.05} & \uline{92.80} & 94.03\tabularnewline
    \midrule 
    \textbf{MV-HetGNN}$_\text{HAN}$ & 59.32 & 59.57 & 14.04 & 14.01 & 93.34 & 93.80 & 79.91 & 85.16 & 80.06 & 83.60\\
    % \midrule 
    \textbf{MV-HetGNN}$_\text{MAGNN}$ & 60.48 & 60.47 & 14.21 & 15.27 & 94.37 & 94.66 & 81.85 & 86.83 & 92.69 & \uline{94.31}\\
    % \midrule 
    \textbf{MV-HetGNN}$_\text{w/o TransE}$ & 60.80 & 60.83 & 13.44 & 14.32 & 95.17 & 95.38 & 83.09 & 87.93 & 92.56 & 93.37\\
    % \midrule 
    \textbf{MV-HetGNN}$_\text{GAT}$ & \textbf{61.87} & \textbf{61.90} & \uline{15.89} & \uline{16.93} & \textbf{95.65} & \textbf{95.93} & \uline{83.83} & \uline{88.68} & \textbf{93.31} & \textbf{95.74}\\
    % \hline
    \midrule 
    \textbf{MV-HetGNN}$_\text{concat}$ & 54.47 & 55.14 & 9.72 & 10.01 & 94.54 & 94.96 & 80.08 & 85.94 & 81.40 & 85.27\\
    % \midrule 
    \textbf{MV-HetGNN}$_\text{mean}$ & 60.43 & 60.35 & 15.18 & 16.10 & 94.27 & 94.70 & 81.77 & 86.95 & 82.23 & 85.95\\
    % \midrule 
    \textbf{MV-HetGNN}$_\text{attn}$ & 60.11 & 60.28 & 12.45 & 14.66 & 94.42 & 94.83 & 81.54 & 86.41 & 92.59 & 93.28\\
    % \midrule
    \textbf{MV-HetGNN}$_\text{w/o ae}$ & 61.08 & 61.15 & 15.33 & 16.28 & 94.82 & 95.09 & 82.17 & 86.59 & 83.35 & 86.02\\
    % \midrule
    \textbf{MV-HetGNN}$_\text{w/o reg}$ & 60.47 & 60.43 & 11.16 & 13.24 & 93.58 & 94.26 & 80.22 & 85.83 & 79.84 & 83.10\\
    % \midrule
    \bottomrule
    \end{tabular}
\end{table*}

\subsection{Study on View-specific Ego Graph Encoder}
\label{Section_Study_VEGE}
% 1. view-specific ego graph encoder v.s. metapath-based neighborhoods and metapath instance
%% instance 已经有了
%% HAN是要做的实验
% 2. impact of modeling the representation of relations the the mapping function between heterogeneous nodes
% 3. 得益于图结构，我们可以更灵活地设计聚合。例如，我们可以使用Attention机制来计算C（可以看一下那边是怎么写的，在这里再重复一遍就好了）。实验效果更好，但是考虑到简洁性，我们以图上的度作为系数。
In this section, we conduct extensive experiments to evaluate the effectiveness of view-specific ego graph encoder.
First of all, we compare the view-specific ego graph encoder with other local structure encoding methods, including metapath-based neighborhoods encoder used in HAN~\cite{HAN} and metapath instance encoder used in MAGNN~\cite{MAGNN}.
Specifically, we replace the view-specific ego graph encoder with the two approaches mentioned above, resulting in two variants, MV-HetGNN$_{\text{HAN}}$ and MV-HetGNN$_{\text{MAGNN}}$.
Second, we evaluate the impact of modeling the representations of relations and the mapping function between heterogeneous nodes by a variant MV-HetGNN$_{\text{w/o\ TransE}}$, which removes the TransE mapping function and aggregates information directly.
Finally, benefiting from modeling the local structure under each semantics as graph structure, we can further improve the performance of view-specific ego graph encoder by setting $\frac{1}{C_i}$ in Eq. (\ref{eq:agg}) learnable.
Specifically, we learn $\frac{1}{C_i}$ by graph attention mechanism~\cite{gat}, resulting in a variant MV-HetGNN$_{\text{GAT}}$.
Due to space limitations, we omit the details of these variants. 
Interested readers can refer to their original papers~\cite{HAN, MAGNN, gat}.
% , HAN~\cite{HAN}, MAGNN~\cite{MAGNN} and GAT~\cite{gat}.

% 现象：View-specific ego graph encoder consistently outperforms approaches in HAN and MAGNN, evaluate the effectiveness of our approaches.
% TransE play a positive rule.
% flexibility
The results are shown in Table \ref{tab:Ablation Study}.
MV-HetGNN consistently outperforms MV-HetGNN$_{\text{HAN}}$ and MV-HetGNN$_{\text{MAGNN}}$, validating the superiority of the view-specific ego graph encoder.
Furthermore, the results of MV-HetGNN$_{\text{w/o\ TransE}}$ indicate that modeling the mapping function plays a positive role in HGs embedding.
Last but not least, MV-HetGNN$_{\text{GAT}}$ outperforms MV-HetGNN in node classification and link prediction tasks, which indicates that the view-specific graph modeling provides us with more flexibility to design models.

\subsection{Study on Auto Multi-view Fusion Layer}
\label{Section_Study_AMF}
% 1. auto multi-view fusion layer v.s. concatenation, mean pooling, and attention
% 3. variants of AMF: 1. w/o hierarchical ae ; 2. w/o orthogonal regularization
In this section, we aim to evaluate the effectiveness of the auto multi-view fusion layer.
First of all, we replace it with other popular metapath fusion methods, including simple concatenation, mean pooling, and the attention mechanism~\cite{HAN, MAGNN, 2020WWWR1, Tree, 2020SCCR2}, resulting in three variants, MV-HetGNN$_{\text{concat}}$, MV-HetGNN$_{\text{mean}}$, and MV-HetGNN$_{\text{attn}}$.
Specifically, the MV-HetGNN$_{\text{concat}}$ simply concatenates the representations from multiple views without any transformation, the MV-HetGNN$_{\text{mean}}$ conducts mean pooling on these representations, and the attention mechanism adopts the implementation in MAGNN~\cite{MAGNN}.
Second, we conduct ablation studies by two variants, MV-HetGNN$_{\text{w/o ae}}$ and MV-HetGNN$_{\text{w/o reg}}$, which remove the hierarchical autoencoder~(replaced with a linear layer) and orthogonal regularization, respectively.

The results are shown in Table \ref{tab:Ablation Study}.
MV-HetGNN consistently outperforms MV-HetGNN$_{\text{concat}}$, MV-HetGNN$_{\text{mean}}$, and MV-HetGNN$_{\text{attn}}$, indicating auto multi-view fusion layer's effectiveness.
Compared with other approaches, MV-HetGNN$_{\text{concat}}$ performs worst. Although it preserves all information from multiple views intuitively, there is much redundant information as discussed in Section \ref{Section_Ortho}, which significantly hinders the performance in downstream tasks.
Furthermore, the MV-HetGNN$_{\text{attn}}$ can not consistently outperform MV-HetGNN$_{\text{mean}}$, especially on the IMDb dataset.
Similar results can also be found in a recent work GIAM~\cite{2021GIAM}, which further shows that the main reason for this phenomenon is overfitting.
% indicates that overfitting is the main reason for this phenomenon.
In addition, the results of MV-HetGNN$_{\text{w/o ae}}$ and MV-HetGNN$_{\text{w/o reg}}$ indicate that both hierarchical autoencoders and orthogonal regularization are critical.
The orthogonal regularization is essential for the success of network optimization, while the auto encoders can further strengthen the performance.
% by providing versatility.
% therefore, the auto multi-view fusion layer can consistently outperform all other popular baselines.

\subsection{Impact of the Length and Number of MetaPaths}
\label{Section_Impact}

\begin{figure}
\centering  %图片全局居中
\setlength{\belowcaptionskip}{-0.2cm}
 \begin{minipage}[h]{\linewidth}
\subfigure[Performance on metapaths of different lengths and semantics.]{
% \label{Fig.sub.1}
%  of view-specific ego graph encoder
 \hspace{-0.15cm}\includegraphics[width=1\textwidth]{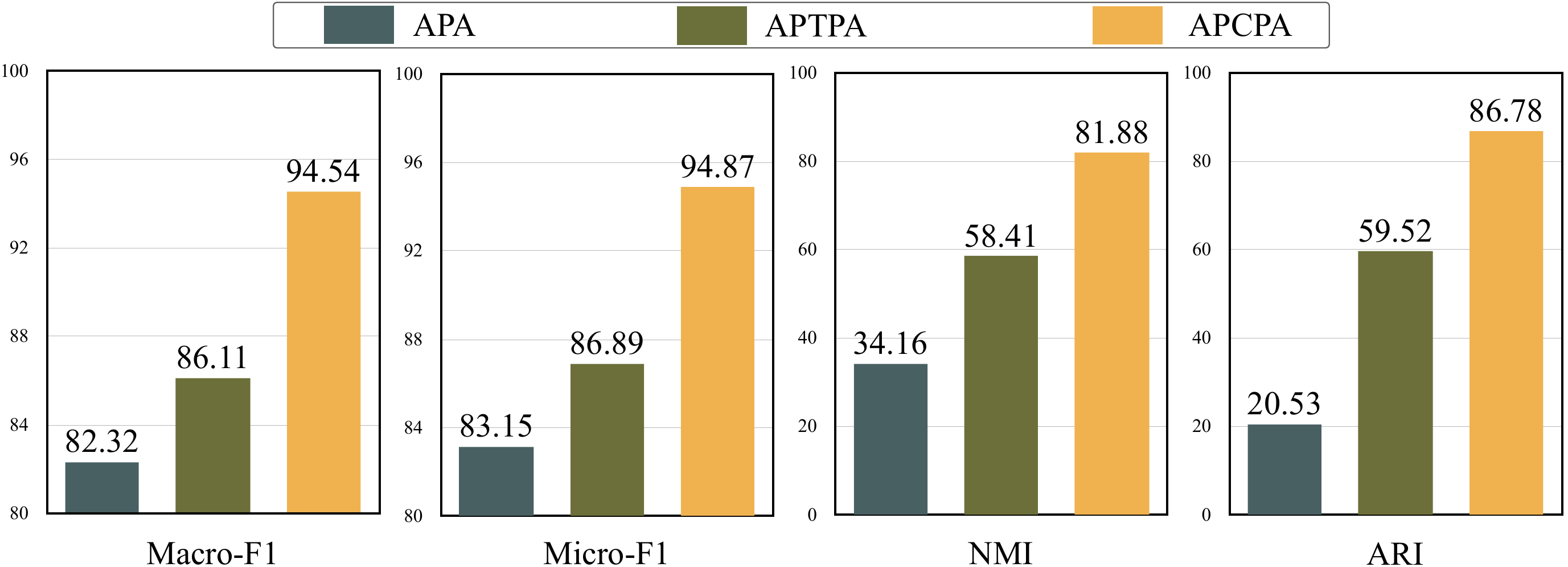}}
\subfigure[Performance on different numbers of metapaths.]{
% \label{Fig.sub.2}
% of MV-HetGNN 
 \hspace{-0.15cm}\includegraphics[width=1\textwidth]{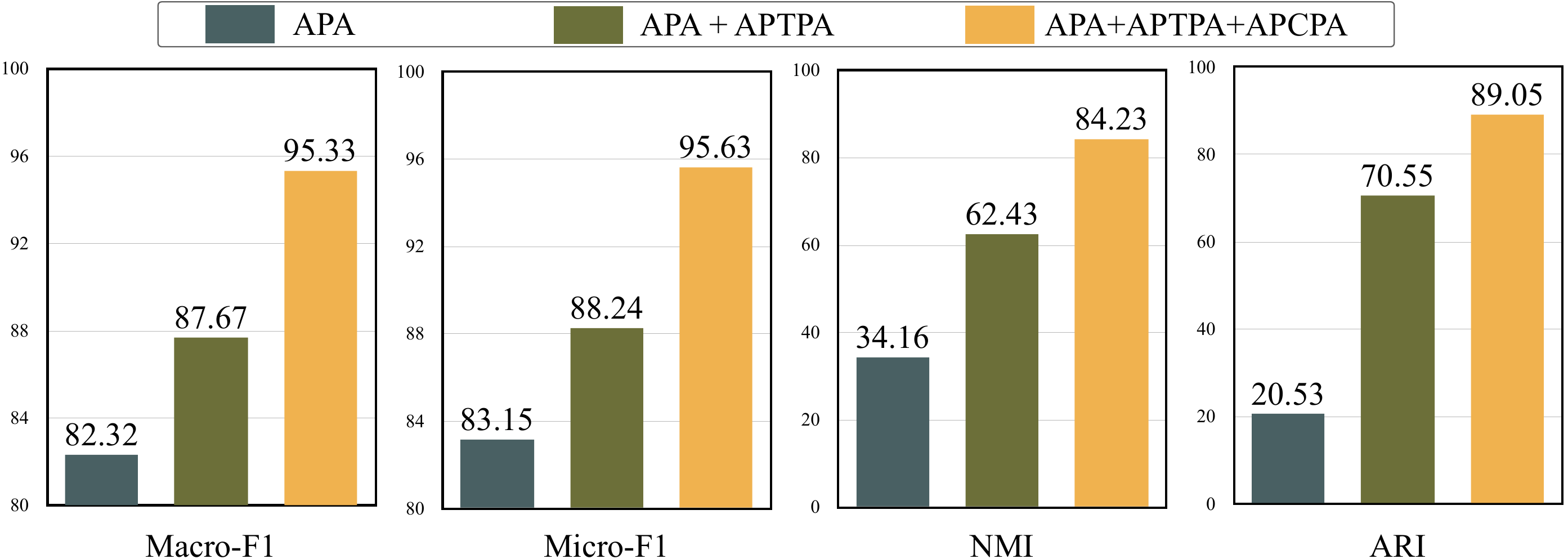}}
\caption{Study on the impact of the length and number of metapaths.}
\label{Fig:Impact of Length and Number of MetaPaths}
  \end{minipage}
\end{figure}

% \begin{figure}[t]
%   \centering
%   \setlength{\abovecaptionskip}{0.1cm}
%   \setlength{\belowcaptionskip}{-0.5cm}
%   \begin{minipage}[h]{\linewidth}
%   \hspace{-0.15cm}\includegraphics[width=0.97\linewidth]{Figures/Introduction_Illustration.pdf}
%   \caption{Examples of traffic flow MTS data and the indistinguishable samples in the spatial and temporal dimension.}
%   \label{Intro}
%   \end{minipage}
% \end{figure}

In this section, we experiment with the DBLP dataset to evaluate the impact of the length and number of metapaths.

First, we compare the performance on three metapaths~(APA, APCPA, and APTPA) individually to evaluate the impact of the length of metapaths.
Specifically, we remove the auto multi-view fusion layer and train MV-HetGNN solely on one of the metapaths. 
Second, we compare the performance on different numbers of metapaths to evaluate the impact of the number of metapaths.
Specifically, we experiment on three groups of metapaths, APA, APA+APTPA, and APA+APTPA+APCPA.
We report the average Macro-F1, average Micro-F1, NMI, and ARI.

% (a) 随着MetaPath长度的上升，性能不一定会变得更差。（b）相比于MetaPath的长度，meta path代表的语义的重要性也是一个很重要的影响下游任务的因素。（换行）（c）In addition，一些自动学习metapath的相关工作的结果显示\cite[ieGNN, GTN]，精炼的MetaPath一般是更好的。这是由于精炼的 MetaPath 更容易产生更明显和有意义的语义。它一般不会过长或者过短，
% 基于过短的MetaPath的局部结构无法容纳足够丰富的结构和特征信息，while基于过长的MetaPath的局部结构引入了过多的噪声信息，他们都无法产生更好的表征。
From Figure \ref{Fig:Impact of Length and Number of MetaPaths}(a), we have two observations.
First, the performance on metapaths APTPA and APCPA outperforms APA, indicating that the performance of view-specific ego graph encoder does not get worse with the increase of metapath length. 
Second, although the lengths of metapath APCPA and APTPA are the same, their performance is quite different. Therefore, compared with the length of metapaths, the semantics behind them are more crucial factors affecting the performance in downstream tasks. 
In addition, some related works~\cite{2021ieGNN, GTN} that can automatically learn metapaths also show that refined metapaths~(\eg 2$\sim$4 order) are generally better. 
Local structures under a too-short metapath cannot include enough structural and feature information, while local structures under a too-long metapath introduce too much noise, and they both cannot be used to produce better node embeddings.

% （a）得益于Auto Multi-View Fusion Layer，随着metapath数量的增加，MV-HetGNN的性能始终在增加，且至少比任意单个metapath下的性能要强。
% （b）我们发现在DBLP数据集上，单APCPA性能，要超过MV-HetGNN_{attn}的性能，which 基于全部三个metapath，并使用流行的attention机制融合不同的metapath。这再次验证了我们的方法的优越性。
As shown in Figure \ref{Fig:Impact of Length and Number of MetaPaths}(b), benefiting from auto multi-view fusion layer, the performance of MV-HetGNN increases consistently with the number of metapaths and is at least as good as any single metapath.
Moreover, we find that view-specific ego graph encoder based on a single metapath APCPA~(Figure \ref{Fig:Impact of Length and Number of MetaPaths}(a)) outperforms the MV-HetGNN$_{attn}$~(Table \ref{tab:Ablation Study}), which integrates all three metapaths by the attention mechanism used in many works~\cite{HAN, MAGNN}.
This phenomenon again validates the superiority of our method.

\color{black}
\subsection{Parameter Sensitivity}
In this section, we conduct experiments to analyze the impacts of two critical hyper-parameters. 

\begin{figure}[h]
\centering  %图片全局居中
\setlength{\abovecaptionskip}{-0.cm}
\setlength{\belowcaptionskip}{-0.5cm}
\subfigure[Dimension of the $\mathbf{H}$]{
% \label{Fig.sub.1}
\includegraphics[width=0.23\textwidth]{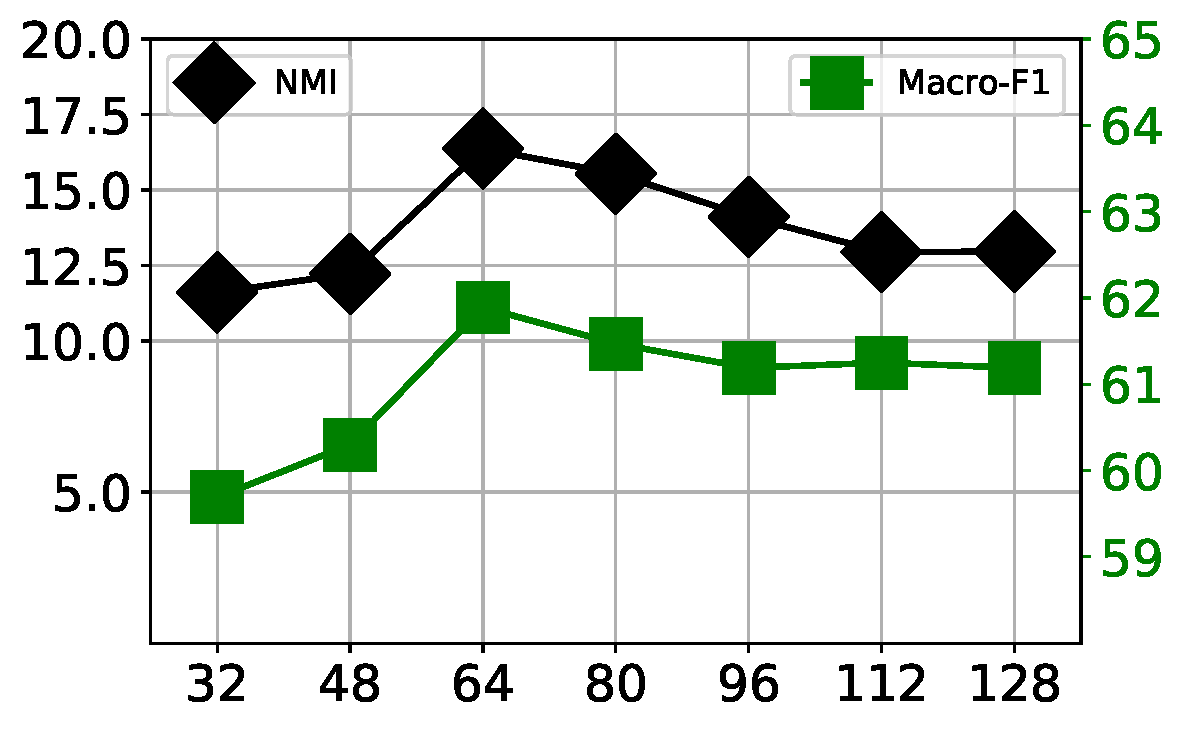}}
\subfigure[Value of $\lambda$ in Eq. (\ref{total_loss})]{
% \label{Fig.sub.2}
\includegraphics[width=0.24\textwidth]{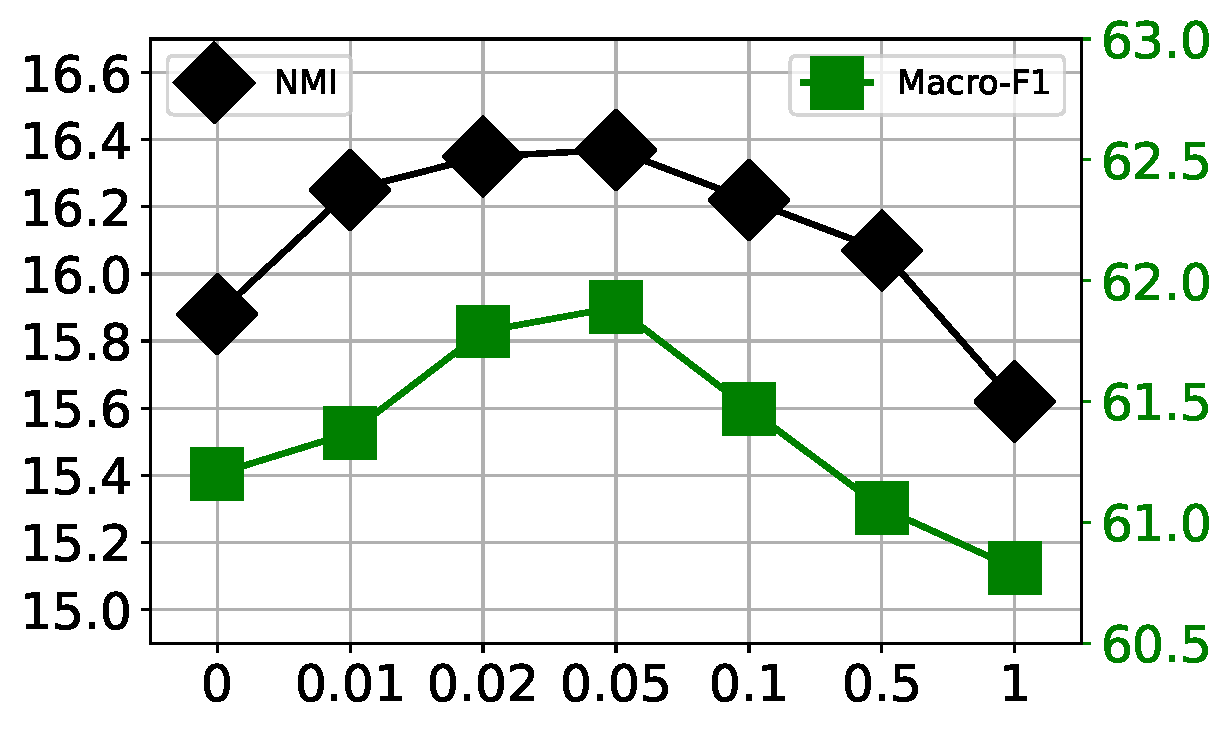}}
\caption{Parameter sensitivity of MV-HetGNN}
\label{Ablation}
\end{figure}

In Figure \ref{Ablation}, 
  we present the results of NMI in node clustering and Macro-F1 in node classification on the IMDb dataset with different parameters.

First, we test sensitivity of the representation feature dimension $d$ of $\mathbf{H}$.
According to the setting in Section \ref{setup}, we additionally set $d^{'}=d^{\frac{M}{2}}=d$.
Note that we only set it for the convenience of the parameter sensitivity test.
The effect of dimension $d$ of the final embedding $\mathbf{H}$ is shown in Figure \ref{Ablation}(a). 
As the embedding dimension increases, the performance will gradually rise to the highest point and then drop slowly. 
This is because the smaller dimension is not enough to encode the heterogeneous feature information and semantic information, 
  while the larger dimensions may introduce redundancy, hindering the model optimization.

Second, we test the sensitivity of $\lambda$ in Eq. (\ref{total_loss}).
Figure \ref{Ablation}(b) reports the effect of $\lambda$.
As discussed in the Section \ref{sec:optimization}, 
  considering various data sets and downstream tasks, 
  too strong versatility constraints~(higher $\lambda$) can harm the model performance.
A relatively small value will improve the performance of MV-HetGNN.
For example, assuming that complementary metapaths that are meaningful to downstream tasks coexist with meaningless metapaths,
  it is beneficial to integrate all the information from the complementary metapaths,
  but it is harmful to forcibly contain information from the meaningless metapaths.

\begin{figure}
\setlength{\abovecaptionskip}{-0.cm}
\setlength{\belowcaptionskip}{-0.4cm}
\centering
\subfigure[Metapath2vec]{
\includegraphics[width=0.225\textwidth]{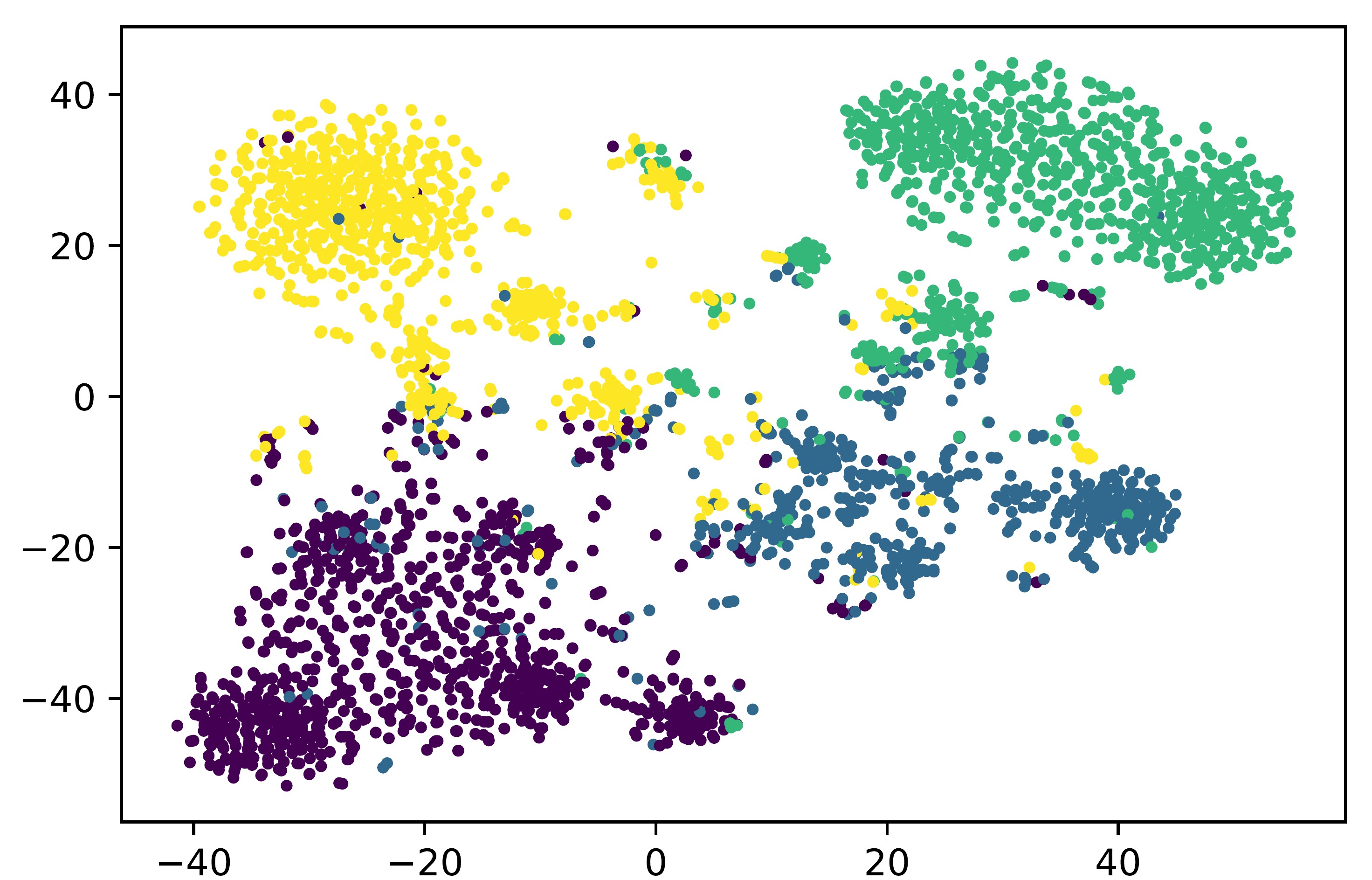}
%\caption{fig1}
}
\vspace{-.1in}
% \quad
\subfigure[GAT]{
\includegraphics[width=0.225\textwidth]{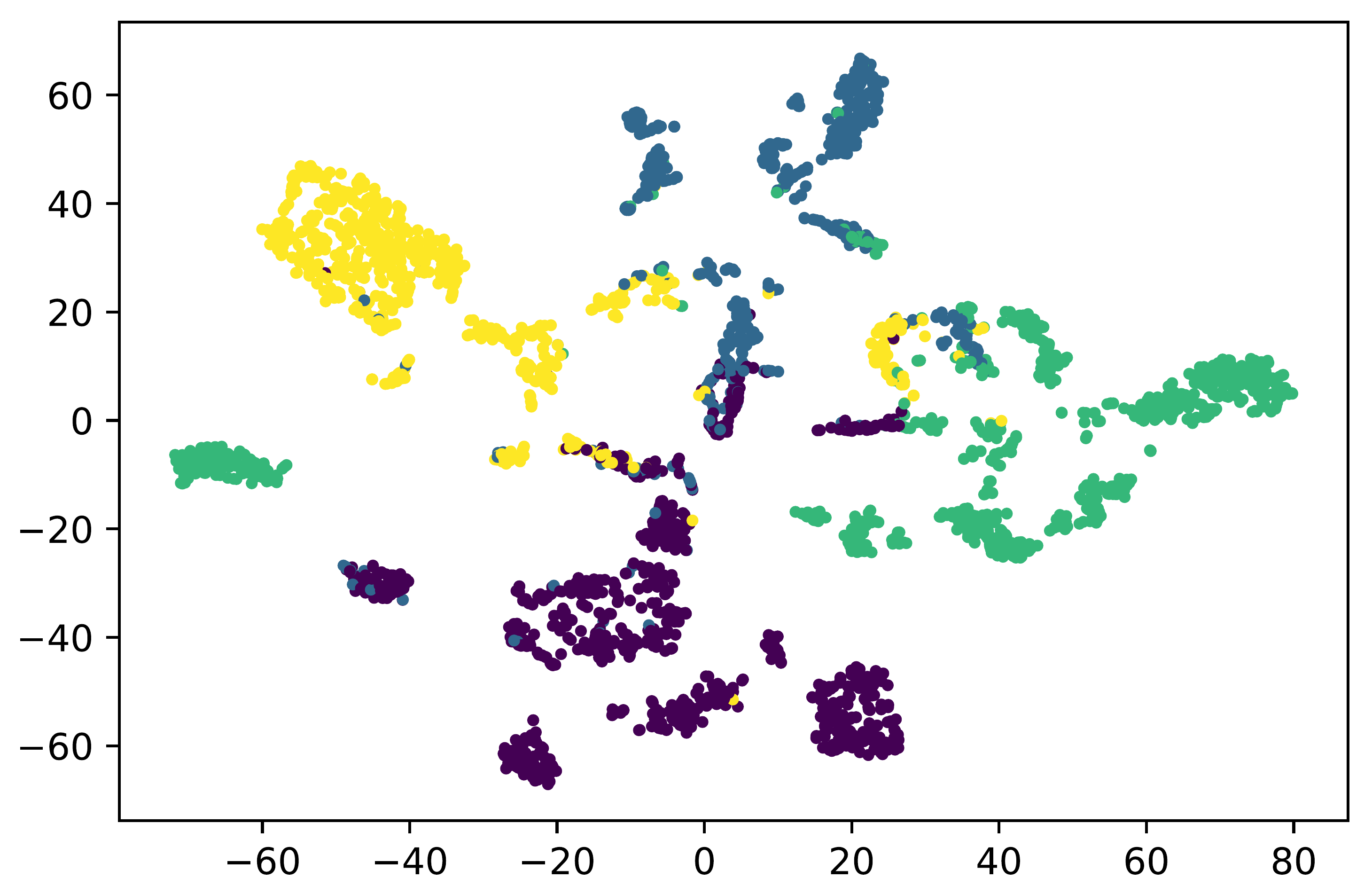}
}

% \quad
\subfigure[HAN]{
\includegraphics[width=0.225\textwidth]{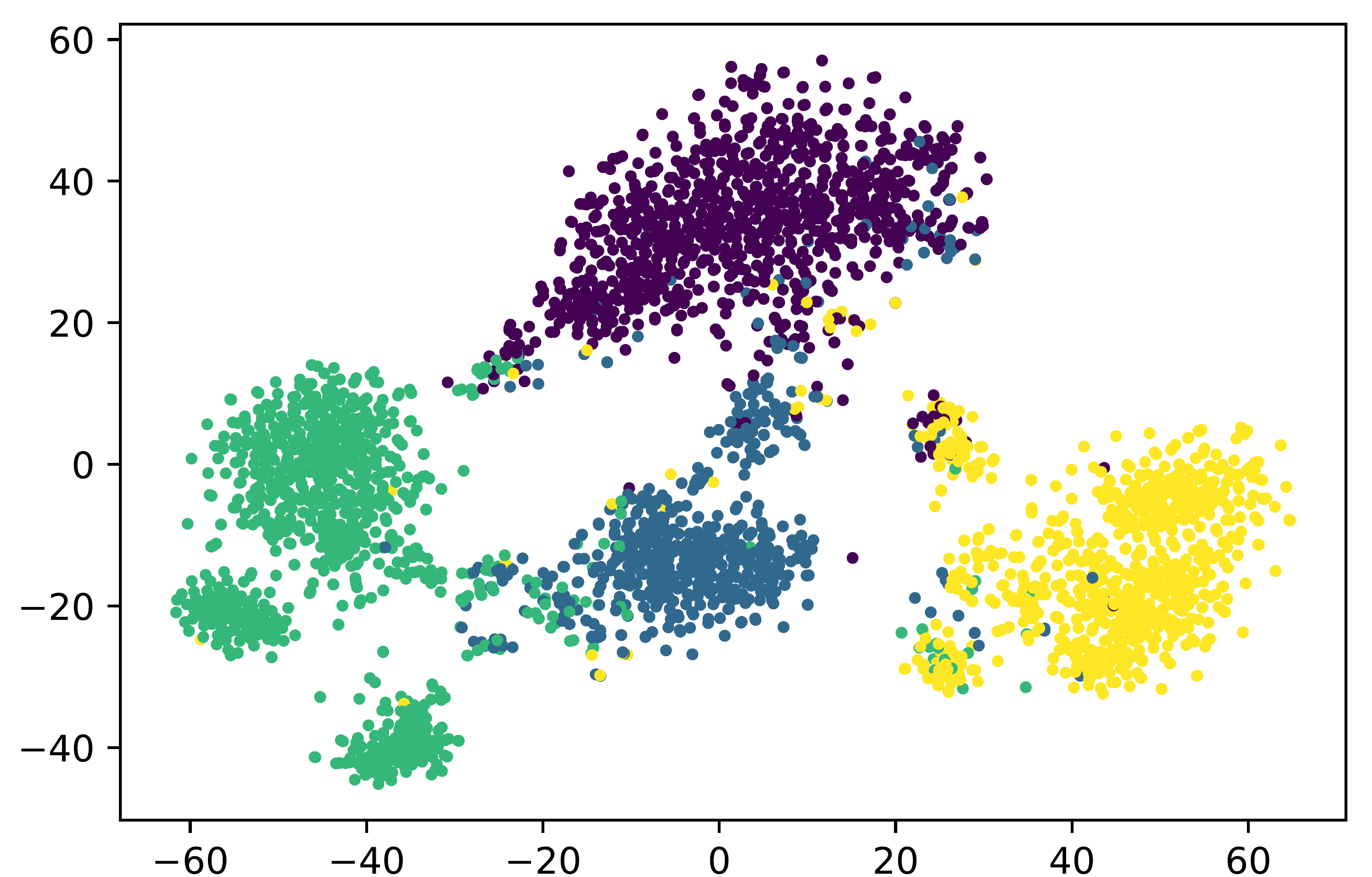}
}
% \quad
\subfigure[MV-HetGNN]{
\includegraphics[width=0.225\textwidth]{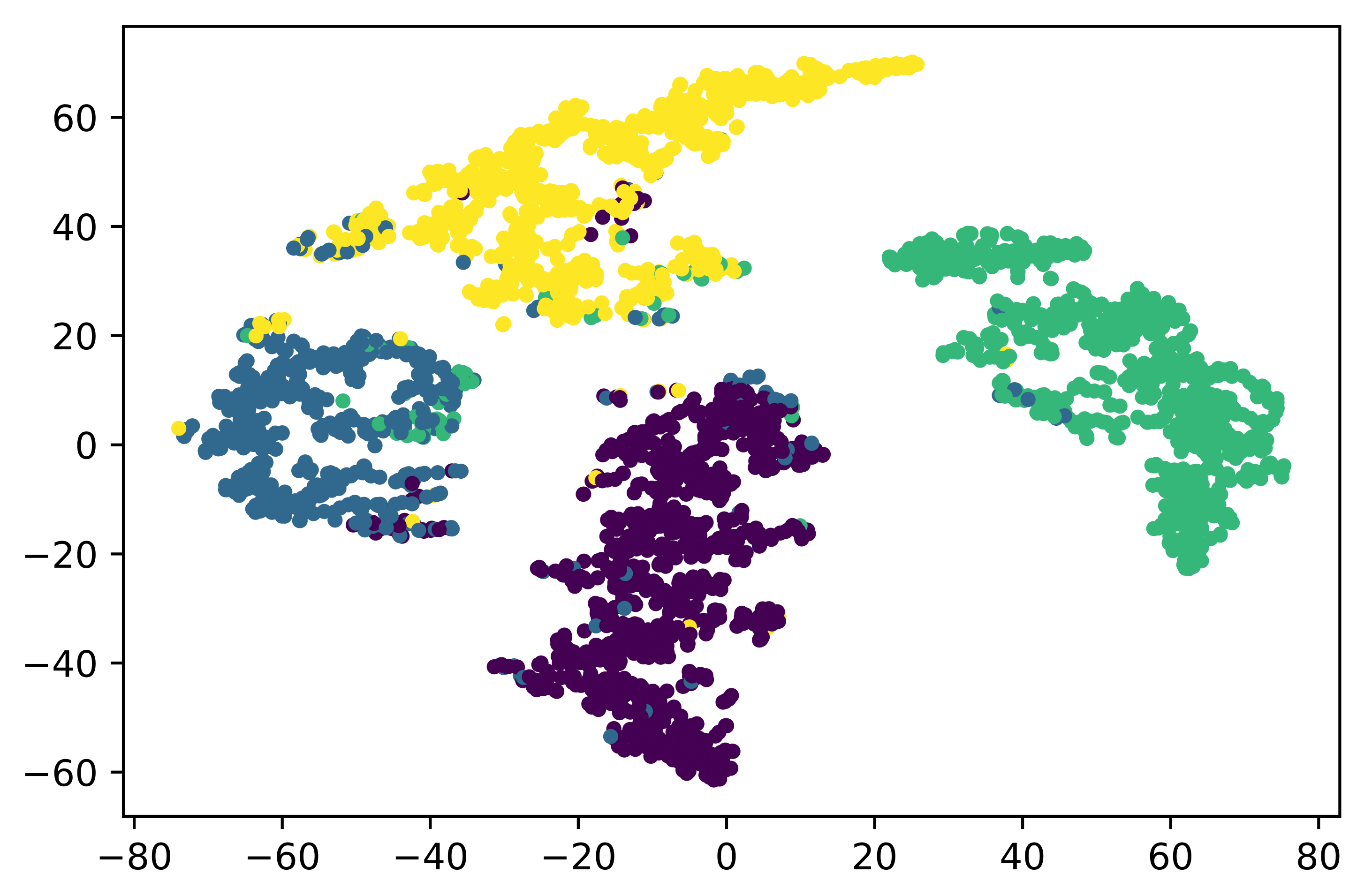}
}
\caption{Visualization of embeddings on DBLP. Each point denotes one author and its color indicates the research area.}
\label{Visualization}
\end{figure}

\subsection{Visualization} 
% \subsubsection{Visualization}~\\
% 1. 同质图模型GAT表现最差：同类节点之间分布不紧密，不同类节点之间混叠。
% 2. 相比于MP2V，HAN利用了多种语义，因此其结果更加致密一些。
% 3. MV-HetGNN是最优秀的，他混叠更少，同类节点之间分布更加致密、不同类节点之间的边界更加清晰。
In this part, we conduct the task of visualization to intuitively compare the embedding results on a low dimensional space. First, we get the embeddings of the nodes in the testing set and then project them into a 2-dimensional space through t-SNE~\cite{t-SNE}.
% Due to the space limit, 
{\color{black}We present a visualization of author node embeddings in DBLP and color the nodes based on their labels.} 
As shown in Figure \ref{Visualization}, GAT performs worst, where nodes of the same type are not closely distributed, and nodes with different types are mixed. 
Benefiting from the use of multiple metapaths to explore comprehensive semantic information, the visualization of HAN performs better than Metapath2vec and GAT.
Further, MV-HetGNN achieves the best visualization performance. 
% The distribution of nodes of the same type is denser, and the boundaries between nodes of different types are clearer compared to all three baselines.
The nodes of the same type are located close to each other, and the nodes from different types are well separated.
% The distribution of nodes of the same type is denser, and the boundaries between nodes of different types are more apparent than all three baselines.

\section{Conclusion}
\label{Section6}
In this paper, 
    we introduce the idea of multi-view representation learning to HGs embedding and propose a \textit{Heterogeneous Graph Neural Network with multi-view representation learning}~(MV-HetGNN).
The complex ego graph in HGs is decomposed into multiple view-specific ego graphs based on different semantics.
Then, we employ the view-specific ego graph encoder to obtain node representation under each view.
In this process, heterogeneity is addressed by learning the representation of relations and modeling the mapping relation between heterogeneous nodes.
Then the auto multi-view fusion layer is developed to integrate the embeddings from diverse views.
Versatile node embeddings with a theoretical guarantee are learned in this module.
We conduct extensive experiments on three real-world datasets, and the results show that the proposed MV-HetGNN significantly outperforms all the baselines on various tasks.

\ifCLASSOPTIONcompsoc
  \section*{Acknowledgments}
\else
  \section*{Acknowledgment}
\fi

This work was supported in part by the National Natural Science Foundation of China under Grant No. 61902376, No. 61902382, and No. 62276110, 
in part by CCF-AFSG Research Fund under Grant No. RF20210005, and in part by the fund of Joint Laboratory of HUST and Pingan Property \& Casualty Research (HPL).
In addition, Zhao Zhang is supported by the China Postdoctoral Science Foundation under Grant No. 2021M703273.
The authors would also like to thank the anonymous reviewers for their comments on improving the quality of this paper.

\ifCLASSOPTIONcaptionsoff
  \newpage
\fi

\bibliographystyle{IEEEtrans}
\normalem
\bibliography{references}

\begin{thebibliography}{10}
\providecommand{\url}[1]{#1}
\csname url@samestyle\endcsname
\providecommand{\newblock}{\relax}
\providecommand{\bibinfo}[2]{#2}
\providecommand{\BIBentrySTDinterwordspacing}{\spaceskip=0pt\relax}
\providecommand{\BIBentryALTinterwordstretchfactor}{4}
\providecommand{\BIBentryALTinterwordspacing}{\spaceskip=\fontdimen2\font plus
\BIBentryALTinterwordstretchfactor\fontdimen3\font minus
  \fontdimen4\font\relax}
\providecommand{\BIBforeignlanguage}[2]{{%
\expandafter\ifx\csname l@#1\endcsname\relax
\typeout{** WARNING: IEEEtran.bst: No hyphenation pattern has been}%
\typeout{** loaded for the language `#1'. Using the pattern for}%
\typeout{** the default language instead.}%
\else
\language=\csname l@#1\endcsname
\fi
#2}}
\providecommand{\BIBdecl}{\relax}
\BIBdecl

\bibitem{social1}
W.~Fan, Y.~Ma, Q.~Li, Y.~He, E.~Zhao, J.~Tang, and D.~Yin, ``Graph neural
  networks for social recommendation,'' in \emph{WWW}, 2019.

\bibitem{social2}
M.~Zhang and Y.~Chen, ``Link prediction based on graph neural networks,'' in
  \emph{NeurIPS}, 2018.

\bibitem{ying2018graph}
R.~Ying, R.~He, K.~Chen, P.~Eksombatchai, W.~L. Hamilton, and J.~Leskovec,
  ``Graph convolutional neural networks for web-scale recommender systems,'' in
  \emph{SIGKDD}, 2018.

\bibitem{wang2020m2grl}
M.~Wang, Y.~Lin, G.~Lin, K.~Yang, and X.-m. Wu, ``M2grl: A multi-task
  multi-view graph representation learning framework for web-scale recommender
  systems,'' in \emph{SIGKDD}, 2020.

\bibitem{graphsage}
W.~Hamilton, Z.~Ying, and J.~Leskovec, ``Inductive representation learning on
  large graphs,'' in \emph{NeurIPS}, 2017.

\bibitem{gcn}
T.~N. Kipf and M.~Welling, ``Semi-supervised classification with graph
  convolutional networks,'' in \emph{ICLR}, 2017.

\bibitem{idgnn}
J.~You, J.~M. Gomes-Selman, R.~Ying, and J.~Leskovec, ``Identity-aware graph
  neural networks,'' in \emph{AAAI}, 2021.

\bibitem{2020methods}
X.~Wang, D.~Bo, C.~Shi, S.~Fan, Y.~Ye, and P.~S. Yu, ``A survey on
  heterogeneous graph embedding: Methods, techniques, applications and
  sources,'' \emph{arXiv preprint arXiv:2011.14867}, 2020.

\bibitem{HAN}
X.~Wang, H.~Ji, C.~Shi, B.~Wang, Y.~Ye, P.~Cui, and P.~S. Yu, ``Heterogeneous
  graph attention network,'' in \emph{WWW}, 2019.

\bibitem{MAGNN}
X.~Fu, J.~Zhang, Z.~Meng, and I.~King, ``Magnn: Metapath aggregated graph
  neural network for heterogeneous graph embedding,'' in \emph{WWW}, 2020.

\bibitem{RGCN}
M.~Schlichtkrull, T.~N. Kipf, P.~Bloem, R.~Van Den~Berg, I.~Titov, and
  M.~Welling, ``Modeling relational data with graph convolutional networks,''
  in \emph{ESWC}, 2018.

\bibitem{HetGNN}
C.~Zhang, D.~Song, C.~Huang, A.~Swami, and N.~V. Chawla, ``Heterogeneous graph
  neural network,'' in \emph{SIGKDD}, 2019.

\bibitem{RelGNN}
X.~Qin, N.~Sheikh, B.~Reinwald, and L.~Wu, ``Relation-aware graph attention
  model with adaptive self-adversarial training,'' in \emph{AAAI}, 2021.

\bibitem{HGT}
Z.~Hu, Y.~Dong, K.~Wang, and Y.~Sun, ``Heterogeneous graph transformer,'' in
  \emph{WWW}, 2020.

\bibitem{2020WWWR1}
Q.~Zhong, Y.~Liu, X.~Ao, B.~Hu, J.~Feng, J.~Tang, and Q.~He, ``Financial
  defaulter detection on online credit payment via multi-view attributed
  heterogeneous information network,'' in \emph{Proceedings of The Web
  Conference 2020}, 2020, pp. 785--795.

\bibitem{2020SCCR2}
F.~Xie, Z.~Cao, Y.~Xu, L.~Chen, and Z.~Zheng, ``Graph neural network and
  multi-view learning based mobile application recommendation in heterogeneous
  graphs,'' in \emph{2020 IEEE International Conference on Services Computing
  (SCC)}.\hskip 1em plus 0.5em minus 0.4em\relax IEEE, 2020, pp. 100--107.

\bibitem{MEIRec}
S.~Fan, J.~Zhu, X.~Han, C.~Shi, L.~Hu, B.~Ma, and Y.~Li, ``Metapath-guided
  heterogeneous graph neural network for intent recommendation,'' in
  \emph{Proceedings of the 25th SIGKDD}, 2019, pp. 2478--2486.

\bibitem{RecoGCN}
F.~Xu, J.~Lian, Z.~Han, Y.~Li, Y.~Xu, and X.~Xie, ``Relation-aware graph
  convolutional networks for agent-initiated social e-commerce
  recommendation,'' in \emph{Proceedings of the 28th CIKM}, 2019, pp. 529--538.

\bibitem{HGSRec}
H.~Ji, J.~Zhu, X.~Wang, C.~Shi, B.~Wang, X.~Tan, Y.~Li, and S.~He, ``Who you
  would like to share with? a study of share recommendation in social
  e-commerce,'' in \emph{AAAI}, 2021.

\bibitem{Tree}
Z.~Qiao, P.~Wang, Y.~Fu, Y.~Du, P.~Wang, and Y.~Zhou, ``Tree structure-aware
  graph representation learning via integrated hierarchical aggregation and
  relational metric learning,'' \emph{arXiv preprint arXiv:2008.10003}, 2020.

\bibitem{TransE}
A.~Bordes, N.~Usunier, A.~Garcia-Duran, J.~Weston, and O.~Yakhnenko,
  ``Translating embeddings for modeling multi-relational data,''
  \emph{NeurIPS}, 2013.

\bibitem{AE2}
C.~Zhang, Y.~Liu, and H.~Fu, ``Ae2-nets: Autoencoder in autoencoder networks,''
  in \emph{Proceedings of the IEEE/CVF Conference on Computer Vision and
  Pattern Recognition}, 2019, pp. 2577--2585.

\bibitem{CPMNets}
C.~Zhang, Y.~Cui, Z.~Han, J.~T. Zhou, H.~Fu, and Q.~Hu, ``Deep partial
  multi-view learning,'' \emph{IEEE TPAMI}, 2020.

\bibitem{firstgcn}
J.~Bruna, W.~Zaremba, A.~Szlam, and Y.~LeCun, ``Spectral networks and locally
  connected networks on graphs,'' in \emph{ICLR}, 2014.

\bibitem{chebnet}
M.~Defferrard, X.~Bresson, and P.~Vandergheynst, ``Convolutional neural
  networks on graphs with fast localized spectral filtering,'' in
  \emph{NeurIPS}, 2016.

\bibitem{gat}
P.~Velickovic, G.~Cucurull, A.~Casanova, A.~Romero, P.~Li{\`{o}}, and
  Y.~Bengio, ``Graph attention networks,'' in \emph{{ICLR}}, 2018.

\bibitem{mpnn}
J.~Gilmer, S.~S. Schoenholz, P.~F. Riley, O.~Vinyals, and G.~E. Dahl, ``Neural
  message passing for quantum chemistry,'' in \emph{ICML}, 2017.

\bibitem{transformer}
A.~Vaswani, N.~Shazeer, N.~Parmar, J.~Uszkoreit, L.~Jones, A.~N. Gomez,
  {\L}.~Kaiser, and I.~Polosukhin, ``Attention is all you need,'' in
  \emph{NeurIPS}, 2017.

\bibitem{amgcn}
X.~Wang, M.~Zhu, D.~Bo, P.~Cui, C.~Shi, and J.~Pei, ``Am-gcn: Adaptive
  multi-channel graph convolutional networks,'' in \emph{SIGKDD}, 2020.

\bibitem{bestgcn}
P.~Li, Y.~Wang, H.~Wang, and J.~Leskovec, ``Distance encoding: Design provably
  more powerful neural networks for graph representation learning,''
  \emph{NeurIPS}, 2020.

\bibitem{wltest}
K.~Xu, W.~Hu, J.~Leskovec, and S.~Jegelka, ``How powerful are graph neural
  networks?'' in \emph{{ICLR}}, 2019.

\bibitem{GATNE}
Y.~Cen, X.~Zou, J.~Zhang, H.~Yang, J.~Zhou, and J.~Tang, ``Representation
  learning for attributed multiplex heterogeneous network,'' in \emph{SIGKDD},
  2019.

\bibitem{zhang2019shne}
C.~Zhang, A.~Swami, and N.~V. Chawla, ``Shne: Representation learning for
  semantic-associated heterogeneous networks,'' in \emph{WSDM}, 2019.

\bibitem{yd1}
C.~Yue, L.~Du, Q.~Fu, W.~Bi, H.~Liu, Y.~Gu, and D.~Yao, ``Htgn-btw:
  Heterogeneous temporal graph network with bi-time-window training strategy
  for temporal link prediction,'' \emph{arXiv preprint arXiv:2202.12713}, 2022.

\bibitem{fu2019metapath}
Y.~Fu, Y.~Xiong, S.~Y. Philip, T.~Tao, and Y.~Zhu, ``Metapath enhanced graph
  attention encoder for hins representation learning,'' in \emph{2019 IEEE
  International Conference on Big Data (Big Data)}, 2019.

\bibitem{yd2}
X.~Chu, X.~Fan, D.~Yao, C.-L. Zhang, J.~Huang, and J.~Bi, ``Noise-aware network
  embedding for multiplex network,'' in \emph{IJCNN}.\hskip 1em plus 0.5em
  minus 0.4em\relax IEEE, 2019, pp. 1--8.

\bibitem{zhu2019relation}
S.~Zhu, C.~Zhou, S.~Pan, X.~Zhu, and B.~Wang, ``Relation structure-aware
  heterogeneous graph neural network,'' in \emph{ICDM}, 2019.

\bibitem{HetSANN}
H.~Hong, H.~Guo, Y.~Lin, X.~Yang, Z.~Li, and J.~Ye, ``An attention-based graph
  neural network for heterogeneous structural learning,'' in \emph{AAAI}, 2020.

\bibitem{comgcn}
S.~Vashishth, S.~Sanyal, V.~Nitin, and P.~P. Talukdar, ``Composition-based
  multi-relational graph convolutional networks,'' in \emph{ICLR}, 2020.

\bibitem{DistMult}
B.~Yang, W.-t. Yih, X.~He, J.~Gao, and L.~Deng, ``Embedding entities and
  relations for learning and inference in knowledge bases,'' \emph{arXiv
  preprint arXiv:1412.6575}, 2014.

\bibitem{ReLU}
V.~Nair and G.~E. Hinton, ``Rectified linear units improve restricted boltzmann
  machines,'' in \emph{ICML}, 2010.

\bibitem{direction}
D.~Marcheggiani and I.~Titov, ``Encoding sentences with graph convolutional
  networks for semantic role labeling,'' in \emph{EMNLP}, 2017.

\bibitem{SAE}
L.~Le, A.~Patterson, and M.~White, ``Supervised autoencoders: Improving
  generalization performance with unsupervised regularizers,'' \emph{Advances
  in neural information processing systems}, vol.~31, pp. 107--117, 2018.

\bibitem{OrthorAE}
C.~Ranjan, \emph{Understanding Deep Learning Application in Rare Event
  Prediction}.\hskip 1em plus 0.5em minus 0.4em\relax Connaissance Publishing,
  2020.

\bibitem{word2vec}
T.~Mikolov, I.~Sutskever, K.~Chen, G.~S. Corrado, and J.~Dean, ``Distributed
  representations of words and phrases and their compositionality,''
  \emph{NeurIPS}, 2013.

\bibitem{DBLP_e}
M.~Ji, Y.~Sun, M.~Danilevsky, J.~Han, and J.~Gao, ``Graph regularized
  transductive classification on heterogeneous information networks,'' in
  \emph{{ECML} {PKDD}}, 2010.

\bibitem{LastFM}
I.~Cantador, P.~Brusilovsky, and T.~Kuflik, ``2nd workshop on information
  heterogeneity and fusion in recommender systems (hetrec 2011),'' in
  \emph{RecSys}, 2011.

\bibitem{metapath2vec}
Y.~Dong, N.~V. Chawla, and A.~Swami, ``metapath2vec: Scalable representation
  learning for heterogeneous networks,'' in \emph{SIGKDD}, 2017.

\bibitem{skip-gram}
T.~Mikolov, K.~Chen, G.~Corrado, and J.~Dean, ``Efficient estimation of word
  representations in vector space,'' in \emph{ICLR}, 2013.

\bibitem{HERec}
C.~Shi, B.~Hu, W.~X. Zhao, and S.~Y. Philip, ``Heterogeneous information
  network embedding for recommendation,'' \emph{TKDE}, 2018.

\bibitem{deepwalk}
B.~Perozzi, R.~Al-Rfou, and S.~Skiena, ``Deepwalk: Online learning of social
  representations,'' in \emph{SIGKDD}, 2014.

\bibitem{2021GIAM}
D.~Jin, Z.~Yu, D.~He, C.~Yang, P.~Yu, and J.~Han, ``Gcn for hin via implicit
  utilization of attention and meta-paths,'' \emph{IEEE TKDE}, 2021.

\bibitem{2021ieGNN}
Y.~Yang, Z.~Guan, J.~Li, W.~Zhao, J.~Cui, and Q.~Wang, ``Interpretable and
  efficient heterogeneous graph convolutional network,'' \emph{IEEE TKDE},
  2021.

\bibitem{GTN}
S.~Yun, M.~Jeong, R.~Kim, J.~Kang, and H.~J. Kim, ``Graph transformer
  networks,'' in \emph{NeurIPS}, 2019.

\bibitem{t-SNE}
L.~Van~der Maaten and G.~Hinton, ``Visualizing data using t-sne.'' \emph{JMLR},
  2008.

\end{thebibliography}

\begin{IEEEbiography}[{\includegraphics[width=1in,height=1.25in,clip,keepaspectratio]{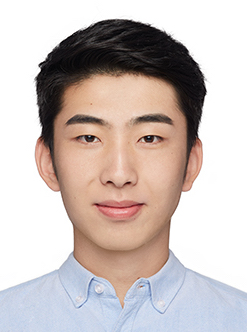}}]
  {Shao  Zezhi} 
  received the B.E. degree from Shandong University, Jinan, China, in 2019.
  He is currently pursuing a Ph.D. degree 
    with the Institute of Computing Technology, Chinese Academy of Sciences, China.
  His research interests include graph and data mining.
\end{IEEEbiography}

\begin{IEEEbiography}[{\includegraphics[width=1in,height=1.25in,clip,keepaspectratio]{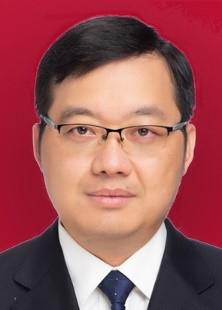}}]
  {Yongjun Xu} is a professor at Institute of Computing Technology, Chinese Academy of Sciences (ICT-CAS) in Beijing, China. He received his B.Eng. and Ph.D. degree in computer communication from Xi'an Institute of Posts \& Telecoms (China) in 2001 and Institute of Computing Technology, Chinese Academy of Sciences, Beijing, China in 2006, respectively. His current research interests include artificial intelligence systems, and big data processing. 
\end{IEEEbiography}

\begin{IEEEbiography}[{\includegraphics[width=1in,height=1.25in,clip,keepaspectratio]{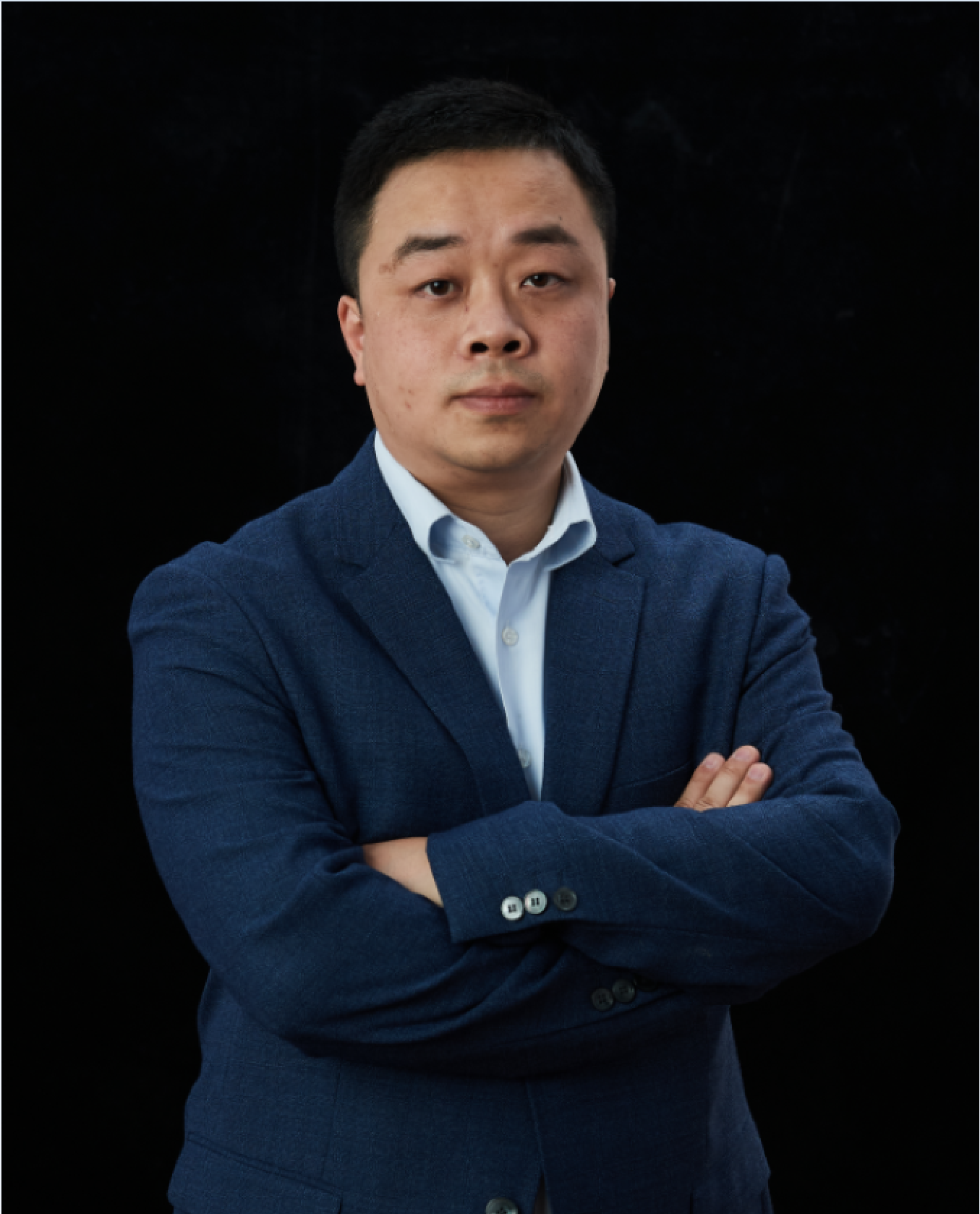}}]
  {Wei  Wei} received the PhD degree from the Huazhong University of Science and Technology, Wuhan, China, in 2012. He is currently an Associate Professor with the Department of Computer of Science and Technology, Huazhong University of Science and Technology. He was a research fellow with Nanyang Technological University, Singapore, and Singapore Management University, Singapore. His current research interests include information retrieval, natural language processing, social computing and recommendation, cross-modal/multimodal computing,  deep learning, machine learning and artificial intelligence.
\end{IEEEbiography}

\begin{IEEEbiography}[{\includegraphics[width=1in,height=1.25in,clip,keepaspectratio]{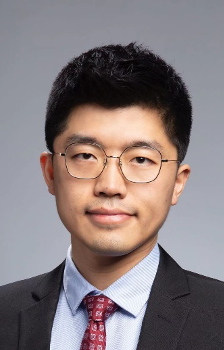}}]
  {Fei Wang}, born in 1988, PhD, associate professor. He received the B.S. degree in computer science from the Beijing Institute of Technology, Beijing, China, in 2011. He received the PhD degree in computer architecture from Institute of Computing Technology, Chinese Academy of Sciences in 2017. From 2017 to 2020, he was a research assistant with the Institute of Technology, Chinese Academy of Sciences. Since 2020, he has been working as associate professor in Institute of Computing Technology, Chinese Academy of Sciences. His main research interest includes spatiotemporal data mining, Information fusion, graph neural networks.
\end{IEEEbiography}

\begin{IEEEbiography}[{\includegraphics[width=1in,height=1.25in,clip,keepaspectratio]{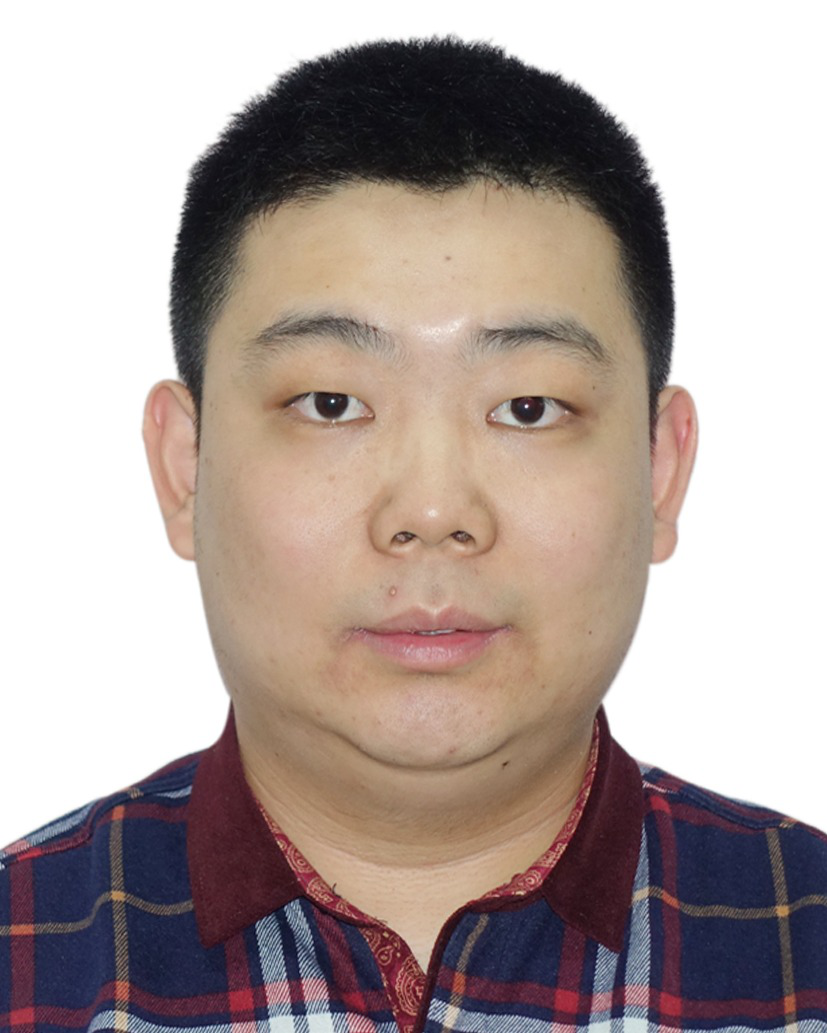}}]
  {Zhao Zhang}
  is a research associate at the Institute of Computing Technology, Chinese Academy of Sciences, Beijing, China. He received the B.E. degree in Computer Science and Technology from the Beijing Institute of Technology (BIT) in 2015, and Ph.D. degree in the Institute of Computing Technology, Chinese Academy of Sciences in 2021. His current research interests include data mining and knowledge graphs.
\end{IEEEbiography}

\begin{IEEEbiography}[{\includegraphics[width=1in,height=1.25in,clip,keepaspectratio]{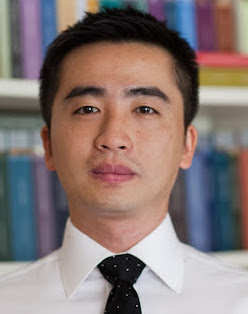}}]
{Feida Zhu} received the B.S. degree in computer science from Fudan University, Shanghai, China, and the Ph.D. degree in computer science from the University of Illinois at Urbana–Champaign, Champaign, IL, USA.
He is an Assistant Professor with the School of Information Systems, Singapore Management University, Singapore. His current research interests include large-scale graph pattern mining and social network analysis, with applications on Web, management information systems, business intelligence, and bioinformatics.
\end{IEEEbiography}

\end{document}